\documentclass{article}

\usepackage[main, final]{neurips_2026}

\usepackage[utf8]{inputenc} 
\usepackage[T1]{fontenc}    
\usepackage{hyperref}       
\usepackage{url}            
\usepackage{booktabs}       
\usepackage{amsfonts}       
\usepackage{nicefrac}       
\usepackage{microtype}      
\usepackage{xcolor}         
\usepackage{graphicx}
\usepackage{xspace}
\usepackage{mathtools} 
\usepackage{amssymb}
\usepackage{amsthm}
\usepackage{bm}
\usepackage{bbm}
\usepackage[mathscr]{eucal}
\usepackage{xfrac}
\usepackage[ruled,noend,vlined]{algorithm2e} 
\usepackage{tabularx}
\usepackage{environ}
\usepackage{caption}
\usepackage{float}
\usepackage{subcaption}

\DeclareMathAlphabet\mathbfcal{OMS}{cmsy}{b}{n}

\newtheoremstyle{example}
  {9pt}
  {9pt}
  {}
  {}
  {\bfseries}
  {.}
  { }
  {}
\theoremstyle{example}

\newtheoremstyle{theorem}
  {9pt}
  {9pt}
  {}
  {}
  {\bfseries}
  {.}
  { }
  {}
\theoremstyle{theorem}
\newtheorem{theorem}{Theorem}

\newtheorem{corollary}{Corollary}
\newtheorem{lemma}{Lemma}
\newtheorem{proposition}{Proposition}

\newtheorem{remark}{Remark}

\makeatletter
\newcommand{\problemtitle}[1]{\gdef\@problemtitle{#1}}
\newcommand{\probleminput}[1]{\gdef\@probleminput{#1}}
\newcommand{\problemquestion}[1]{\gdef\@problemquestion{#1}}
\NewEnviron{problem}{
  \problemtitle{}\probleminput{}\problemquestion{}
  \BODY
  \par\addvspace{.5\baselineskip}
  \noindent
  \begin{tabularx}{\textwidth}{@{\hspace{.25\parindent}} l X c}
    \multicolumn{2}{@{\hspace{.25\parindent}}l}{\@problemtitle} \\
    \textbf{Input:} & \@probleminput \\
    \textbf{Question:} & \@problemquestion
  \end{tabularx}
  \par\addvspace{.5\baselineskip}
}
\makeatother

\AddToHook{cmd/appendix/before}{%
  \setcounter{equation}{0}%
  \setcounter{lemma}{0}
  
  \setcounter{theorem}{0}%
  \setcounter{corollary}{0}%
  \setcounter{proposition}{0}%
  \setcounter{remark}{0}%
}

\renewcommand{\mathcal}{\CMcal}

\renewcommand{\emptyset}{\varnothing}

\renewcommand{\vec}[1]{\bm{#1}}

\newcommand{\Probability}{\mathbb P}
\newcommand{\Expectation}{\mathbb E}

\newcommand{\Ind}{\mathbb I}

\newcommand{\opt}{\mathrm{opt}}

\newcommand{\ClassNP}{\ensuremath{\textsf{NP}}\xspace}

\newcommand{\ClassNPPP}{\ensuremath{\smash{\textsf{NP}^\textsf{PP}}}\xspace}
\newcommand{\ClassSigmaTwoP}{\ensuremath{\smash{\mathsf{\Sigma}_2^{\textsf{p}}}}\xspace}
\newcommand{\ClassWTwo}{\ensuremath{\textsf{W}[2]}\xspace}

\title{Learning Global Probabilistic Explanations}

\author{
  Frédéric Koriche \\
  Univ.~Artois, CNRS\\ 
  Centre de Recherche en Informatique de Lens (CRIL), France\\
  \texttt{frederic.koriche@cril.fr} \\
  \And
	Louenas Bounia \\
  Université Sorbonne Paris Nord, CNRS\\
  Laboratoire d'Informatique de Paris Nord (LIPN), UMR 7030, France\\
  \texttt{bounia@lipn.univ-paris13.fr} \\
}

\begin{document}

\maketitle


\begin{abstract}
	Interpreting the predictions of complex black-box classifiers remains a central challenge in explainable artificial intelligence.
	While local explanations clarify individual predictions, there is a significant need for global probabilistic explanations that capture a model's overall behavior across the input distribution.
	Such an explanation is a small subset $K$ of features, whose relevance is measured by the expected probability, over a random input $\vec x$, that the classifier assigns the same label to $\vec x$ and to a random input agreeing with $\vec x$ on $K$.
	In this study, we investigate the task of identifying maximally relevant global explanations under a cardinality constraint, focusing on product distributions with full support on discrete feature spaces.
	We introduce a spectral explainer that leverages membership queries to the black-box classifier and employs Fourier-analytic, junta-learning methods to produce probably approximately correct (PAC) explanations.
	Our algorithm is fixed-parameter tractable with respect to the explanation size limit, maximum feature cardinality, and desired accuracy.
	Experiments on synthetic and real-world datasets confirm the theoretical guarantees and show that the spectral explainer outperforms a greedy baseline as the explanation size increases.
\end{abstract}


\section{Introduction}
\label{sec:introduction}

Classification has long been the cornerstone of machine learning research~\cite{ShalevShwartz.Book.2014},
evolving from early linear models and decision trees to modern deep neural networks, ensemble methods,
and graph-based architectures~\cite{Sayed.Book.2022}.
While maximizing predictive accuracy was once the primary objective, the integration of machine learning into high-stakes,
real-world applications~\cite{Lipton.Queue.2018} has made interpretability and transparency essential.
In these contexts, strong performance alone is insufficient---stakeholders demand explanations that are clear, understandable, and well-justified.
This imperative has catalyzed rapid advances in \emph{explainable artificial intelligence (XAI)},
a field dedicated to making complex models transparent and accessible to human users~\cite{Burkart.JAIR.2021,Molnar.Book.2025}.

A central question in XAI is to explain \emph{why} a classifier $f$ assigns a specific label $f(\vec{x})$ to an input instance $\vec{x}$.
Among the many explanation types developed to address this need~\cite{Dasgupta.ICML.2022,Ribeiro.KDD.2016}, decision rules or \emph{anchors}~\cite{Ribeiro.AAAI.2018} stand out for their conceptual clarity.
When $\vec x$ is described by a set of $d$ interpretable features, an anchor can be represented by a subset $K \subseteq [d]$ such that altering features outside $K$ is not likely to affect the outcome.
Notably, if the restriction $\vec{x}_K$ of $\vec{x}$ to $K$ suffices to determine $f(\vec{x})$ with certainty,
$K$ is known as an \emph{abductive explanation}~\cite{Cooper.AI.2023,Ignatiev.AAAI.2019} or \emph{sufficient reason}~\cite{Barcelo.NeurIPS.2020,Darwiche.ECAI.2020}.

Despite their strong theoretical foundations, abductive explanations are frequently too large to be practical~\cite{Ignatiev.CP.2020, Ignatiev.AAAI.2019}.
This limitation stems from human cognition: psychological research consistently finds that people can process
only a small number of interacting variables---typically four or five---before cognitive overload occurs~\cite{Halford.BBS.1998,JohnsonLaird.PNAS.2010}.
Moreover, the complexity of an explanation depends not only on the number of features $k$, but also on their domain cardinality $c$:
since the number of configurations grows as $c^k$, the demand on working memory rises exponentially.
Consequently, empirical XAI research emphasizes the need for concise explanations to achieve genuine interpretability~\cite{Lage.HCOMP.2019, Narayanan.arXiv.2018}.

Motivated by these cognitive limitations,
recent advances in XAI have shifted toward \emph{probabilistic (abductive) explanations},
which aim to strike a balance between accuracy and interpretability~\cite{Blanc.NeurIPS.2021,Izza.JAR.2023,Waldchen.JAIR.2021}.
Within this framework, the explanatory power of a feature subset $K$ is measured by its \emph{local relevance}:
the probability that a classifier $f$ assigns the same label to a random instance $\vec{z}$ as to a reference $\vec{x}$, provided $\vec{x}$ and $\vec{z}$ agree on all features in $K$.
Given a user-specified size limit $k$ and a distribution $\mathcal{D}$ over instances, the objective is to find a subset $K$ of size at most $k$ that maximizes:
\begin{align}
	\label{eq:local_relevance}
	\Probability_{\vec z \sim \mathcal D(\cdot \mid \vec z_K = \vec x_K)}\left[f(\vec x) = f(\vec z)\right].
\end{align}
For example, consider a loan approval model $f$ that evaluates applicants based on \texttt{Age}, \texttt{Income}, and \texttt{Debt-to-Income}.
Suppose an applicant $\vec x$ is denied a loan: although their debt-to-income ratio is low, their income falls below the model's threshold.
In this setting, \texttt{Income} acts as a relevant local explanation: for any applicant $\vec{z}$ with the same income, the model will likely reach the same decision, regardless of age.
Thus, knowing income alone suffices to predict $\vec x$'s outcome with high probability.

While prior work has focused on the \emph{local} perspective---explaining individual predictions---recent research increasingly pursues a more ambitious \emph{global} perspective,
aiming to capture a model's behavior across the entire input space~\cite{Bassan.ICML.2024,Bassan.ICLR.2026}.
In this setting, the feature subset $K$ serves as a \emph{junta}~\cite{Blum.COLT.03,Blum.AI.1997}, partitioning the input space into regions defined by all possible assignments to $K$.
With a size constraint $k$ and a distribution $\mathcal{D}$, the goal is to find a subset $K$ of size at most $k$ that maximizes:
\begin{align}
	\label{eq:global_relevance}
	\Probability_{\vec x \sim \mathcal D,\; \vec z \sim \mathcal D(\cdot \mid \vec z_K = \vec x_K)}\left[f(\vec x) = f(\vec z)\right].
\end{align}
This global formulation assesses how consistently the model behaves within each region defined by $K$.
Returning to the loan example, $K = \{\texttt{Income}\}$ might be adequate for one applicant, but may show low global relevance if the model's outcomes depend on feature interactions.
In contrast, a pair such as $K = \{\texttt{Income}, \texttt{Debt\mbox{-}to\mbox{-}Income}\}$ can partition applicants into groups---such as ``High Income, Low Debt''---where decisions are consistently positive.
Maximizing \eqref{eq:global_relevance} thus uncovers the feature combinations that drive reliable model behavior across the input distribution.

In essence, \eqref{eq:global_relevance} is simply the average of \eqref{eq:local_relevance}
when $\vec x$ is no longer supplied by the user but drawn from $\mathcal D$.
Yet, this shift fundamentally reshapes the computational landscape for generating explanations.

Recent work by Bassan et al.~\cite{Bassan.ICLR.2026} investigates this distinction by considering the problem of minimizing
the size of an explanation subject to a threshold constraint on the relevance error.
They show that, under an empirical product distribution $\mathcal D$ with constant support size,
the global explanation problem is approximable within a constant factor via a greedy algorithm, whereas the local version remains inapproximable.
However, in high-dimensional settings, such empirical distributions cover only a negligible fraction of the input space.
Consequently, a global explanation $K$ tailored to this sparse support may fail to capture the true logic of $f$,
lacking the look-ahead necessary to detect complex feature interactions that emerge only when a substantial portion of
the input space is observed.

By contrast, our results address product distributions that have full support over discrete input spaces, where each feature has cardinality at most $c$.
In this high-support regime, the problem of maximizing~\eqref{eq:local_relevance} subject to a size constraint $k$ is highly intractable
for neural networks~\cite{Waldchen.JAIR.2021}, and remains hard for decision trees~\cite{Arenas.NeurIPS.2022}.
Further evidence shows that, under standard complexity assumptions, minimizing the size of a highly relevant local explanation
is not approximable in polynomial time~\cite{Kozachinskiy.arXiv.2023, Subercaseaux.AAAI.2025, Waldchen.JAIR.2021},
nor is it fixed-parameter tractable in terms of $c$ and $k$~\cite{Ordyniak.IJCAI.2023}.

Despite this local intractability, we demonstrate that near-optimal global explanations can be computed in fixed-parameter tractable time---even when the classifier $f$ is accessed only as a black box.
Our framework provides a structural characterization of optimal explanations for \eqref{eq:global_relevance} via the Fourier spectrum of $f$ for product distributions over discrete domains~\cite{ODonnell.Book.2014}.
Based on this framework, we introduce a \emph{spectral explainer} that combines an extended version of the Kushilevitz--Mansour (KM) procedure~\cite{Kushilevitz.SICOMP.1993}
for identifying significant Fourier coefficients with the Gopalan--Kalai--Klivans (GKK) pruning strategy~\cite{Gopalan.STOC.2008} for efficient subset selection.
Given any accuracy parameter $\epsilon$ and confidence parameter $\delta$, our algorithm returns, with probability at least $1-\delta$, a feature subset $K$ of size at most $k$ whose global relevance is within $\epsilon$ of optimal.
The algorithm is fixed-parameter tractable with respect to $c$, $k$ and $1/\epsilon$, with a runtime polynomial in $d$, $(ck/\epsilon)^k$, and $\ln(1/\delta)$.

To complement our theoretical findings, we evaluate the spectral explainer on both synthetic and real-world datasets. When the optimal explanation is computable, its relevance closely matches the predicted accuracy, and its query complexity scales with $d$, $k$, and $\epsilon$ as our analysis suggests. We further benchmark it against a greedy baseline inspired by~\cite{Bassan.ICLR.2026}, which lacks guarantees for full-support product distributions. On neural networks, the spectral explainer more frequently finds the optimal explanation, especially as $k$ increases, reflecting the Fourier spectrum's ability to reveal feature interactions missed by greedy selection.


\section{Related Work}
\label{sec:related_work}

We say that a subset $K$ is a \emph{$k$-size $\tau$-relevant} local explanation with respect to a reference instance $\vec{x}$,
a classifier $f$, and a distribution $\mathcal{D}$ if $K$ contains at most $k$ features and its local relevance \eqref{eq:local_relevance} is at least $\tau$.
Analogous terminology applies to global probabilistic explanations by replacing \eqref{eq:local_relevance} with \eqref{eq:global_relevance}.

In the local setting, common choices for $\mathcal{D}$ include neighborhood distributions centered at $\vec{x}$~\cite{Ribeiro.AAAI.2018}
and the uniform distribution over the input space~\cite{Izza.JAR.2023, Koriche.ECML.2024}.
Notably, under the uniform distribution, deciding whether a $k$-size $\tau$-relevant explanation exists for a given $\vec{x}$ is computationally challenging:
the problem is \ClassNPPP-hard for neural networks~\cite{Waldchen.JAIR.2021} and \ClassNP-hard for decision trees~\cite{Arenas.NeurIPS.2022}.
Even for the special case $\tau = 1$, corresponding to finding $k$-size abductive explanations,
the task is \ClassSigmaTwoP-hard for random forests~\cite{Audemard.AAAI.2022} and neural networks~\cite{Barcelo.NeurIPS.2020}.
The prospects for fixed-parameter tractability and approximability remain bleak. Specifically,
finding cardinality-minimal $\tau$-relevant local explanations is \ClassWTwo-hard with respect to $k$,
even when $\tau = 1$ and $f$ is a monotone neural network~\cite{Ordyniak.IJCAI.2023}.
Furthermore, the problem is \ClassNP-hard to approximate within any factor of $d^{1 - \alpha}$,
for any $\alpha > 0$, when $f$ is a neural network~\cite{Waldchen.JAIR.2021} or a decision tree~\cite{Kozachinskiy.arXiv.2023}.
Positive results are limited to relaxed versions of the problem: for linear functions~\cite{Subercaseaux.AAAI.2025},
and for arbitrary black-box classifiers under the uniform distribution, at the cost of larger and less precise explanations~\cite{Blanc.NeurIPS.2021}.

The global perspective, as analyzed by Bassan et al.~\cite{Bassan.ICML.2024,Bassan.ICLR.2026}, is markedly different from the local setting.
Notably, the set function \eqref{eq:global_relevance} is monotone for any distribution $\mathcal{D}$, and supermodular when $\mathcal{D}$ is a product distribution.
As a result, when $\mathcal{D}$ is an empirical product distribution supported on a fixed dataset, a greedy forward selection algorithm approximates
cardinality-minimal $\tau$-relevant explanations within a factor that depends on the curvature of \eqref{eq:global_relevance} and the dataset size.

The dual problem---maximizing the relevance of $k$-size explanations---closely relates to the well-studied task of learning \emph{$k$-juntas}, which are functions depending on at most $k$ out of $d$ variables~\cite{Blum.COLT.03,Blum.AI.1997}. In its agnostic version, the objective is to find a $k$-junta $h$ that maximizes the accuracy $\Probability_{\vec{z} \sim \mathcal{D}}[h(\vec{z}) = f(\vec{z})]$ for a given Boolean function $f$. Learning $k$-juntas from random examples under the uniform distribution is notoriously hard, even in the realizable case: the best known algorithms run in time $d^{\alpha k}$ for some constant $\alpha < 1$~\cite{Mossel.STOC.2003,Valiant.JACM.2015}, and the problem is closely tied to learning noisy parities~\cite{Feldman.SIAM.2009}; see also~\cite{Arpe.TCS.2007} for juntas under attribute noise. With membership queries, Fourier-analytic methods such as the KM algorithm~\cite{Kushilevitz.SICOMP.1993} make the problem tractable: in particular, $k$-juntas are properly and agnostically learnable under the uniform distribution~\cite{Gopalan.STOC.2008}. Although conceptually related, $k$-junta learning results cannot be directly transferred to global explainability, since their objectives---accuracy versus relevance---differ, and global explanations correspond to partitioning functions rather than Boolean ones. Finally, the problem of computing global probabilistic explanations shares similarities with \emph{knowledge distillation}, which seeks to learn a global, interpretable surrogate model $h$ of the black-box classifier $f$ (see, e.g.,~\cite{Burkart.JAIR.2021} for a survey). Yet again, the goals diverge: surrogate models aim to \emph{recover} the classifier by maximizing accuracy (or fidelity),
while global explanations seek to \emph{determine} the classifier's predictions by maximizing relevance.



\section{A Spectral Learning Framework}
\label{sec:framework}

The classifiers under consideration in this study are functions mapping data from a discrete, finite space $\vec{\Omega} = \Omega_1 \times \cdots \times \Omega_d$ to labels in $\{-1, +1\}$.
Each index $i \in [d] = \{1, \ldots, d\}$ specifies an interpretable feature taking values in the alphabet $\Omega_i = \{0, \ldots, c_i - 1\}$.
Let $c = \max_{i \in [d]} c_i$ denote the largest alphabet size.
A product distribution $\mathcal D$ over $\vec{\Omega}$ is defined by $\mathcal D = \mathcal D_1 \times \cdots \times \mathcal D_d$,
where each $\mathcal D_i$ is a probability distribution on $\Omega_i$.
Throughout this work, we assume all marginal distributions have \emph{full support}, meaning that $\mathcal D_i(a) > 0$ for all $i \in [d]$ and $a \in \Omega_i$.

\subsection{Fourier Bases}

The core of our framework is the construction of an orthonormal basis of functions with respect to the underlying distribution~\cite{ODonnell.Book.2014}.
Focusing on a single domain $\Omega = \{0, \ldots, c-1\}$ with distribution $\mathcal{D}$, let $\mathcal{L}^2(\Omega, \mathcal{D})$
denote the Hilbert space with the inner product $\langle f, g \rangle_{\mathcal{D}} = \Expectation_{x \sim \mathcal{D}}[f(x)g(x)]$.
An orthonormal basis consists of $c$ functions $\{\phi_0, \phi_1, \ldots, \phi_{c-1}\}$ satisfying $\langle \phi_j, \phi_k \rangle_{\mathcal{D}} = \Ind[j = k]$.
Since $\langle 1, 1 \rangle_{\mathcal{D}} = 1$, the constant function can always serve as the first basis function.

\begin{lemma}
	\label{lem:local_basis}
	For any distribution $\mathcal{D}$ with full support over an alphabet $\Omega$ of size $c$,
	an orthonormal basis for $\mathcal{L}^2(\Omega, \mathcal{D})$ with $\phi_0 = 1$ can be computed in $O(c^2)$ time using the Stieltjes procedure~\cite{Gautschi.Book.2004}.
\end{lemma}

The Stieltjes procedure treats the values $0, \ldots, c-1$ as distinct real nodes; since any choice of distinct nodes yields a valid basis,
the procedure applies equally to ordinal and categorical features.

For each feature $i \in [d]$, let $\{\phi_{i,0}, \ldots, \phi_{i,c_i-1}\}$ be an orthonormal basis of $\mathcal{L}^2(\Omega_i, \mathcal{D}_i)$ with $\phi_{i,0} = 1$.
By taking the tensor product of these local bases, we derive a global basis for the space $\mathcal{L}^2(\vec{\Omega}, \mathcal{D})$.
Let $\mathcal{S} = \smash{\textstyle \prod_{i=1}^d \{0, 1, \ldots, c_i - 1\}}$ denote the set of multi-indices, referred to as \emph{coordinates}.
The \emph{support} of a coordinate $\vec{s} \in \mathcal{S}$ is given by $\mathrm{supp}(\vec{s}) = \{i \in [d] : s_i \neq 0\}$,
and its \emph{global basis function} is defined as $\phi_{\vec{s}}(\vec{x}) = \smash{\textstyle \prod_{i=1}^d \phi_{i,s_i}(x_i)}$.

\begin{lemma}
	\label{lem:global_basis}
	Let $\mathcal{D}$ be a product distribution with full support over $\vec{\Omega}$.
	The set $\{\phi_{\vec{s}} : \vec{s} \in \mathcal{S}\}$ forms a complete orthonormal basis for $\mathcal{L}^2(\vec{\Omega}, \mathcal{D})$.
	As a result, any function $f : \vec{\Omega} \to \mathbb{R}$ admits a unique Fourier expansion:
	\begin{align*}
		f(\vec{x}) = \sum_{\vec{s} \in \mathcal{S}} \hat{f}(\vec{s}) \phi_{\vec{s}}(\vec{x})
		\quad \text{where} \quad \hat{f}(\vec{s}) = \langle f, \phi_{\vec{s}} \rangle_{\mathcal{D}} = \Expectation_{\vec{x} \sim \mathcal{D}}[f(\vec{x}) \phi_{\vec{s}}(\vec{x})].
	\end{align*}
\end{lemma}

For any $p \in [1, \infty)$, the functional $p$-norm and the spectral $p$-norm of a function $f$ are defined as:
\begin{align*}
	\|f\|_p = \left(\Expectation_{\vec{x} \sim \mathcal{D}}[|f(\vec{x})|^p]\right)^{1/p} \quad \text{and} \quad L_p(f) = \left(\sum_{\vec{s} \in \mathcal{S}} |\hat{f}(\vec{s})|^p\right)^{1/p}.
\end{align*}
We also use $L_\infty(f) = \max_{\vec{s} \in \mathcal{S}} |\hat{f}(\vec{s})|$ and $L_0(f) = |\mathrm{supp}(f)|$,
where $\mathrm{supp}(f)$ is the spectral support of $f$, that is, the set of coordinates $\vec{s}$ for which $\hat{f}(\vec{s}) \neq 0$.
As a corollary of Lemma~\ref{lem:global_basis}, both the Plancherel theorem,
$\langle f, g \rangle_{\mathcal{D}} = \sum_{\vec{s} \in \mathcal{S}} \hat{f}(\vec{s}) \hat{g}(\vec{s})$,
and Parseval's identity, $\|f\|_2 = L_2(f)$, hold in this general setting. In particular, $\|f\|_2^2 = 1$ when $f$ takes values in $\{-1, +1\}$.

The \emph{total energy} $W(f)$ of a function $f : \vec{\Omega} \to \mathbb{R}$ is given by $W(f) = L_2(f)^2$. More generally, the \emph{energy} $W_{\mathcal{T}}(f)$ of $f$ on a subset $\mathcal{T} \subseteq \mathcal{S}$
is the restriction of $W(f)$ to the coordinates in $\mathcal T$:
\begin{align*}
	W_{\mathcal{T}}(f) = \sum_{\vec{s} \in \mathcal{T}} \hat{f}(\vec{s})^2.
\end{align*}

\subsection{The Integer Kushilevitz--Mansour Algorithm}
\label{sec:IKM}

Building upon the original Kushilevitz--Mansour (KM) algorithm~\cite{Kushilevitz.SICOMP.1993},
which identifies large Fourier coefficients of Boolean functions on the uniform hypercube,
we introduce an extended version tailored for product distributions over general discrete domains.

Given a coordinate $\vec{s} \in \mathcal{S}$, let $\vec{s}_{p:q}$ denote the sequence $(s_p, s_{p+1}, \ldots, s_q)$ for integers $1 \leq p \leq q \leq d$.
The $p$-prefix of $\vec{s}$ comprises its first $p$ indices, with the $0$-prefix defined as the empty sequence $\vec{\lambda}$.
For a prefix $\vec{\sigma}$ of length $p$, we write $\phi_{\vec{\sigma}}(\vec{x}_{1:p}) = \prod_{i=1}^p \phi_{i,\sigma_i}(x_i)$, with $\phi_{\vec{\lambda}} = 1$.
The \emph{weight} of a function $f$ relative to $\vec{\sigma}$ is the energy concentrated on the prefix cone
$\mathcal{C}_{\vec{\sigma}} = \{\vec{s} \in \mathcal{S} : \vec{s}_{1:p} = \vec{\sigma}\}$. Hereafter, we write $W_{\vec{\sigma}}$
instead of $W_{\mathcal{C}_{\vec{\sigma}}}(f)$ when the context is clear.
As in the Boolean case, this weight can be expressed as an expectation over correlated pairs of inputs:
\begin{align}
	\label{eq:weight_identity}
	W_{\vec{\sigma}} = \Expectation_{\vec{u}, \vec{u}' \sim \mathcal{D}_{1:p}, \vec{v} \sim \mathcal{D}_{p+1:d}}\big[f(\vec{u}, \vec{v})\, \phi_{\vec{\sigma}}(\vec{u})\, f(\vec{u}', \vec{v})\, \phi_{\vec{\sigma}}(\vec{u}')\big],
\end{align}
Estimating~\eqref{eq:weight_identity} requires evaluating $f$ on pairs of inputs that share the same suffix $\vec{v}$,
which is where membership queries are needed.

As shown in Algorithm~\ref{alg:ikm}, the Integer Kushilevitz--Mansour (IKM) algorithm recursively searches for prefixes
$\vec{\sigma}$ whose estimated weight $\hat{W}_{\vec{\sigma}}$ is at least $\theta^2/2$.
If such a prefix has length $d$, the corresponding Fourier coefficient is estimated and added to the hypothesis.
Otherwise, the search continues by extending the prefix with all possible values for the next index.

To ensure the efficiency of the IKM algorithm across non-uniform distributions,
we employ the \emph{Median-of-Means} (MoM) estimator~\cite{Lugosi.FCM.2019}, that provides sub-Gaussian concentration guarantees based solely on variance bounds. This matters here because basis functions may take large values: $|\phi_{i,j}|$ can be as large as $1/\sqrt{\min_a \mathcal{D}_i(a)}$, so range-based bounds such as Hoeffding's inequality would deteriorate with the smallest marginal probabilities, whereas variances remain bounded.

\begin{lemma}
	\label{lem:estimation}
	Let $f: \vec{\Omega} \to \{-1,+1\}$ be a classifier and $\mathcal{D}$ a product distribution with full support.
	For any prefix $\vec{\sigma}$ and coordinate $\vec{s}$, the random variables averaged by \textsc{Weight} and \textsc{Coefficient}
	have expectations $W_{\vec{\sigma}}$ and $\hat{f}(\vec{s})$, respectively, and variance at most $1$.
	Consequently, with $b \geq 8 \ln(1/\delta_0)$ blocks of $n \geq 64/\theta^4$ samples each, every call to \textsc{Weight} or \textsc{Coefficient}
	returns an estimate within $\theta^2/4$ of its target with probability at least $1 - \delta_0$.
\end{lemma}

\begin{algorithm}[t!]
	\DontPrintSemicolon
	\SetKwFunction{FIKM}{IKM}
	\SetKwFunction{FSearch}{Search}
	\SetKwFunction{FWeight}{Weight}
	\SetKwFunction{FCoefficient}{Coefficient}
	\SetKwProg{Fn}{Function}{:}{}
	\caption{Integer Kushilevitz--Mansour (IKM) Algorithm}
	\label{alg:ikm}

	\KwIn{Membership oracle for $f$, product distribution $\mathcal{D}$, threshold $\theta$, confidence $\delta$}
	\KwOut{A sparse hypothesis $g: \vec{\Omega} \to \mathbb{R}$}
	\rule{\textwidth}{0.4pt}

	\begin{minipage}[t]{0.47\textwidth}
		\Fn{\FIKM{$f, \theta, \delta$}}
		{
			$g(\vec x) \leftarrow 0$, $c \leftarrow \max_i c_i$\;
			$\delta_0 \leftarrow \theta^2 \delta / (8dc)$\;
			$b \leftarrow \lceil 8 \ln(1/\delta_0) \rceil$, $n \leftarrow \lceil 64/\theta^4 \rceil$\;
			\FSearch{$\vec{\lambda}$}\;
			\Return $g(\vec x)$\;
		}
		\BlankLine
		\Fn{\FSearch{$\vec{\sigma}$}}
		{
			$\hat{W}_{\vec{\sigma}} \leftarrow \FWeight(f, \vec{\sigma})$\;
			\If{$\hat{W}_{\vec{\sigma}} \geq \theta^2/2$}
			{
				\If{$\mathrm{size}(\vec{\sigma}) = d$}
				{
					$\vec{s} \leftarrow \vec{\sigma}$\;
					$\hat{g}(\vec{s}) \leftarrow \FCoefficient(f, \vec{s})$\;
					$g(\vec x) \leftarrow g(\vec x) + \hat{g}(\vec{s}) \phi_{\vec{s}}(\vec x)$\;
				}
				\Else
				{
					$p \leftarrow \mathrm{size}(\vec{\sigma})$\;
					\For{$j = 0$ \KwTo $c_{p+1}-1$}
					{
						\FSearch{$(\vec{\sigma}, j)$}\;
					}
				}
			}
		}
		\BlankLine
	\end{minipage}
	\hfill
	\begin{minipage}[t]{0.48\textwidth}
		\Fn{\FWeight{$f, \vec{\sigma}$}}
		{
			$p \leftarrow \mathrm{size}(\vec{\sigma})$\;
			\For{$j = 1$ \KwTo $b$}{
				Sample $\vec{U}, \vec{U}' \sim \mathcal{D}_{1:p}^n$ and $\vec{V} \sim \mathcal{D}_{p+1:d}^n$\;
				\For{$i = 1$ \KwTo $n$}{
					$r_i \leftarrow f(\vec{U}_i, \vec{V}_i)\, \phi_{\vec{\sigma}}(\vec{U}_i)$\;
					$r'_i \leftarrow f(\vec{U}'_i, \vec{V}_i)\, \phi_{\vec{\sigma}}(\vec{U}'_i)$\;
				}
				$\bar{W}_j \leftarrow \frac{1}{n} \sum_{i=1}^n r_i r'_i$\;
			}
			\Return $\mathrm{median}(\bar{W}_1, \ldots, \bar{W}_{b})$\;
		}
		\BlankLine
		\Fn{\FCoefficient{$f, \vec{s}$}}{
			\For{$j = 1$ \KwTo $b$}{
				Sample $\vec{Z} \sim \mathcal{D}^n$\;
				\For{$i = 1$ \KwTo $n$}{
					$w_i \leftarrow f(\vec{Z}_i)\, \phi_{\vec{s}}(\vec{Z}_i)$\;
				}
				$\bar{C}_j \leftarrow \frac{1}{n} \sum_{i=1}^n w_i$\;
			}
			\Return $\mathrm{median}(\bar{C}_1, \ldots, \bar{C}_{b})$\;
		}
	\end{minipage}
	\BlankLine
\end{algorithm}

The logical structure of the search ensures that IKM recovers all significant spectral energy while producing a sparse hypothesis.
Proposition~\ref{app:prop:logic} in the Appendix establishes the completeness and soundness of this strategy:
when all estimates are accurate, the search retains every prefix whose cone contains a coefficient with $\hat{f}(\vec{s})^2 \geq \theta^2$,
and only prefixes with weight at least $\theta^2/4$.
This leads to our main convergence result:

\begin{theorem}
	\label{thm:convergence}
	Let $f: \vec{\Omega} \to \{-1,+1\}$ be a classifier and $\mathcal{D}$ a product distribution with full support and maximum alphabet size $c$.
	Assume that each membership query to $f$ and each draw from a marginal $\mathcal{D}_i$ take $O(1)$ time.
	Then, for any threshold $\theta \in (0,1]$ and confidence level $\delta \in (0,1)$, the IKM algorithm makes
	$O\big(\frac{dc}{\theta^6} \ln \frac{dc}{\theta \delta}\big)$ membership queries, runs in time $O\big(dc^2 + \frac{d^2 c}{\theta^6} \ln \frac{dc}{\theta \delta}\big)$,
	and returns a hypothesis $g$ such that, with probability at least $1 - \delta$:
	\begin{align*}
		L_{\infty}(f - g) \leq \theta \quad \text{and} \quad L_0(g) \leq \frac{4}{\theta^2}.
	\end{align*}
\end{theorem}

A key aspect of the IKM algorithm's efficiency is its handling of the alphabet size $c$.
Although the coordinate space $\mathcal{S}$ can contain up to $c^d$ coordinates, the runtime remains polynomial in $d$ and $c$.
This tractability arises from spectral pruning: prefix cones at a given depth are disjoint, so by Parseval's identity,
at most $4/\theta^2$ prefixes of weight at least $\theta^2/4$ can be retained at each depth, independently of $c$.
Each retained prefix spawns at most $c$ children, so at most $4c/\theta^2$ prefixes are examined at each depth,
and at most $1 + 4dc/\theta^2$ in total, which explains why the number of queries is nearly linear in $c$.
Moreover, the prefixes of a given depth can all be estimated on a shared batch of samples:
the analysis is unchanged, and the number of queries drops to $(2d+1)\,b\,n = O\big(\frac{d}{\theta^4} \ln \frac{dc}{\theta\delta}\big)$,
independently of the number of examined prefixes (see Remark~\ref{app:rem:shared} in Appendix~\ref{app:theory}).


\section{A Spectral Explanation Algorithm}
\label{sec:algorithm}

In this section, we present the foundations of our explanation method.
We begin by establishing a structural identity, revealing that the global relevance of any feature subset is precisely quantified by its Fourier energy.
Leveraging this characterization, we introduce a spectral explainer that integrates the IKM algorithm with a pruning strategy,
inspired by~\cite{Gopalan.STOC.2008}, to identify near-optimal explanations.

\subsection{Spectral Characterization of Global Relevance}

Given an instance $\vec{x} \in \vec{\Omega}$ and a subset $K \subseteq [d]$, we denote by $\vec{x}_K$ the restriction of $\vec{x}$ to the features in $K$.
We say that two instances $\vec{x}$ and $\vec{z}$ \emph{agree} on $K$ if $\vec{x}_K = \vec{z}_K$.
For a classifier $f: \vec{\Omega} \to \{-1, +1\}$ and a distribution $\mathcal{D}$ over $\vec{\Omega}$, the \emph{global relevance error} of $K$ is defined as:
\begin{align}
	\label{eq:global_relevance_error}
	\mathsf{err}_{f,\mathcal{D}}(K) = \Probability_{\vec{x} \sim \mathcal{D},\; \vec{z} \sim \mathcal{D}(\cdot \mid \vec{z}_K = \vec{x}_K)}\left[f(\vec{x}) \neq f(\vec{z})\right],
\end{align}
that is, one minus the global relevance~\eqref{eq:global_relevance}.
In line with Bassan et al.~\cite{Bassan.ICLR.2026}, this error is the average, over $\vec{x} \sim \mathcal{D}$, of the local relevance error of $K$ at $\vec{x}$.
The global relevance of $K$ can be characterized entirely in terms of the Fourier spectrum of $f$. Specifically, let $f_K$ denote the \emph{projection} of $f$ onto the feature subset $K$, defined as:
\begin{align*}
	f_K(\vec{x}) = \Expectation_{\vec{z} \sim \mathcal{D}}[f(\vec{z}) \mid \vec{z}_K = \vec{x}_K] = \sum_{\vec{s} \in \mathcal{S}_K} \hat{f}(\vec{s}) \phi_{\vec{s}}(\vec{x}),
\end{align*}
where $\mathcal{S}_K = \{\vec{s} \in \mathcal{S} : \mathrm{supp}(\vec{s}) \subseteq K\}$ is the set of coordinates supported on $K$.
The second equality holds because, under a product distribution, $\Expectation[\phi_{\vec{s}}(\vec{z}) \mid \vec{z}_K = \vec{x}_K]$ equals $\phi_{\vec{s}}(\vec{x})$
if $\mathrm{supp}(\vec{s}) \subseteq K$, and $0$ otherwise, since $\phi_{i,s_i}$ has zero mean whenever $s_i \neq 0$.
By Parseval's identity, the total energy of $f_K$ is the energy of $f$ on $\mathcal{S}_K$, that is, $W(f_K) = W_{\mathcal{S}_K}(f)$.

\begin{lemma}
	\label{lem:characterization}
	For any classifier $f: \vec{\Omega} \to \{-1, +1\}$, any product distribution $\mathcal{D}$ with full support on $\vec{\Omega}$, and any feature subset $K$, the global relevance error of $K$ satisfies:
	\begin{align*}
		\mathsf{err}_{f, \mathcal{D}}(K) = \frac{1}{2}\left(1 - W(f_K)\right).
	\end{align*}
\end{lemma}

\begin{proof}
	Let $\vec{x} \sim \mathcal{D}$ and $\vec{z} \sim \mathcal{D}(\cdot \mid \vec{z}_K = \vec{x}_K)$. Since $f$ takes values in $\{-1, +1\}$, $\Ind[f(\vec{x}) \neq f(\vec{z})] = (1 - f(\vec{x})f(\vec{z}))/2$.
	Conditioning on $\vec{x}$ gives $\Expectation[f(\vec{z}) \mid \vec{x}] = f_K(\vec{x})$, so by Plancherel's theorem,
	$\Expectation[f(\vec{x})f(\vec{z})] = \langle f, f_K \rangle_{\mathcal{D}} = \sum_{\vec{s} \in \mathcal{S}_K} \hat{f}(\vec{s})^2 = W(f_K)$.
\end{proof}

Equivalently, the global relevance of $K$ is $(1 + W(f_K))/2$, so finding a maximally relevant explanation of size at most $k$
amounts to finding a subset $K$ with $|K| \leq k$ that maximizes the energy $W(f_K)$.
Note that, when $c_i > 2$ for some $i$, the local bases are not unique, so individual Fourier coefficients depend on the chosen bases.
The global relevance of $K$, however, does not: $W(f_K)$ is the squared spectral $2$-norm of the conditional expectation $f_K$, which is defined without reference to any basis.

\subsection{Spectral Recovery, Feature Pruning, and Subset Selection}

By carefully calibrating the precision of the IKM phase, we ensure that the spectral energy of the hypothesis $g$ serves as a reliable proxy for the energy of $f$
across all possible explanations of size $k$. The following lemma is based on the observation that the projection $f_K$ has at most $|\mathcal{S}_K| = \prod_{i \in K} c_i \leq c^k$ coefficients.

\begin{lemma}
	\label{lem:recovery}
	Let $f: \vec{\Omega} \to \{-1, +1\}$ be a classifier and $g: \vec{\Omega} \to \mathbb{R}$ such that $L_{\infty}(f - g) \leq \epsilon / c^{k/2}$ for some $\epsilon \in (0,1)$.
	Then, for any subset $K \subseteq [d]$ with $|K| \leq k$, the difference between their projected energies satisfies $|W(f_K) - W(g_K)| \leq 3\epsilon$.
\end{lemma}

While Theorem~\ref{thm:convergence} guarantees that the hypothesis $g$ returned by the IKM algorithm is sparse, this sparsity is strictly ``spectral'';
that is, $L_0(g)$ is small. The coordinates in $\mathrm{supp}(g)$ might collectively involve a large number of distinct variables,
which would still require an exhaustive search over $\binom{d}{k}$ possible subsets to find the optimal $K$.
To achieve fixed-parameter tractability, it is essential to prune the feature space to a compact candidate set $I$ of variables that drive the model's behavior.
To this end, we define the \emph{restricted influence} of a feature $i$ of degree $k$, which aggregates the energy over all coordinates involving $i$ up to support size $k$:
\begin{align*}
	\mathrm{Inf}_i^{k}[g] = W_{\mathcal R_i^{k}}(g) = \sum_{\vec{s} \in \mathcal R_i^k} \hat{g}(\vec{s})^2 \quad \text{where} \quad \mathcal R_i^{k} = \{\vec{s} \in \mathcal{S} : s_i \neq 0, |\mathrm{supp}(\vec{s})| \leq k\}.
\end{align*}
The following lemma establishes that features with low restricted influence can be safely disregarded.

\begin{lemma}
	\label{lem:pruning}
	For a function $g: \vec{\Omega} \to \mathbb{R}$ and $\epsilon > 0$, let $I = \{i \in [d] : \mathrm{Inf}_i^{k}[g] \geq \epsilon/k\}$ be the set of influential features.
	Then for any $K \subseteq [d]$ with $|K| \leq k$, it holds that $W(g_K) \leq W(g_{K \cap I}) + \epsilon$.
\end{lemma}

\begin{algorithm}[t]
	\SetArgSty{textrm}
	\DontPrintSemicolon
	\caption{Spectral Explainer (\textsc{SE})}
	\label{alg:se}
	\SetKw{Input}{Input:}
	\BlankLine
	\Input{Membership query access to classifier $f$, product distribution $\mathcal{D}$, size limit $k$, accuracy $\epsilon$, confidence $\delta$.}\;
	\BlankLine

	Set $\epsilon' \leftarrow 2\epsilon/7$\;

	\tcp{Spectral Recovery}
	Run \textsc{IKM} on $f$ with threshold $\theta = \epsilon' / c^{k/2}$ and confidence $\delta$ to obtain $g$\;

	\tcp{Feature Pruning}
	Identify the influence set $I = \{i \in [d] : \mathrm{Inf}_i^{k}[g] \geq \epsilon'/k\}$\;

	\tcp{Subset Selection}
	Find $K \subseteq I$ with $|K| \leq k$ that maximizes the projected energy $W(g_K)$\;
	\BlankLine
	\Return $K$\;
	\BlankLine
\end{algorithm}

Based on these preparatory results, Algorithm~\ref{alg:se} implements the spectral approach to global explainability:
it first applies the IKM algorithm to obtain a spectral hypothesis $g$ that approximates $f$,
then prunes the feature space to a candidate set $I$ of influential variables,
and finally performs an exhaustive search within this reduced set.
Both the influences and the projected energies are computed directly from the coefficients of $g$,
since $W(g_K) = \sum \{\hat{g}(\vec{s})^2 : \vec{s} \in \mathrm{supp}(g),\ \mathrm{supp}(\vec{s}) \subseteq K\}$.
The following theorem establishes that this algorithm is fixed-parameter tractable in $k$, $c$, and $1/\epsilon$,
and achieves near-optimal relevance with high probability.

\begin{theorem}
	\label{thm:explainability}
	There exists an algorithm such that, for any $d$-dimensional feature space $\vec{\Omega}$ with maximum alphabet size $c$,
	any binary classifier $f$ over $\vec{\Omega}$, and any product distribution $\mathcal{D}$ with full support over $\vec{\Omega}$,
	given any size constraint $k \leq d$, parameters $\epsilon, \delta \in (0,1)$, and membership query access to $f$,
	the algorithm returns a subset $K \subseteq [d]$ of size at most $k$ satisfying, with probability at least $1 - \delta$:
	\begin{align*}
		\mathsf{err}_{f, \mathcal{D}}(K) \leq \min_{K'\subseteq [d], |K'| \leq k} \mathsf{err}_{f, \mathcal{D}}(K') + \epsilon.
	\end{align*}
	The algorithm is fixed-parameter tractable in $k$, $c$, and $1/\epsilon$, with a runtime polynomial in $d$, $(ck/\epsilon)^k$, and $\ln(1/\delta)$.
\end{theorem}

\begin{proof}[Sketch of proof]
	Let $K$ be returned by Algorithm~\ref{alg:se}, and $K^*$ be optimal. By Lemma~\ref{lem:characterization}, it suffices to show that $W(f_{K^*}) - W(f_K) \leq 2\epsilon$.
	With probability at least $1 - \delta$, IKM returns $g$ with $L_\infty(f - g) \leq \theta$ (Theorem~\ref{thm:convergence}).
	Then, Lemma~\ref{lem:recovery} changes the energy of $K^*$ and of $K$ by at most $3\epsilon'$ each when passing from $f$ to $g$,
	Lemma~\ref{lem:pruning} loses at most $\epsilon'$ when restricting $K^*$ to $I$, and $K$ maximizes $W(g_{K'})$ over all 
	subsets $K' \subseteq I$ of size at most $k$:
	hence $W(f_{K^*}) - W(f_K) \leq 7\epsilon' = 2\epsilon$.

	For the runtime, Theorem~\ref{thm:convergence} with $\theta = 2\epsilon/(7c^{k/2})$ shows that IKM is polynomial in $d$, $c^k/\epsilon$, and $\ln(1/\delta)$; the key point is the size of $I$.
	By soundness, every $\vec{s} \in \mathrm{supp}(g)$ satisfies $|\hat{f}(\vec{s})| \geq \theta/2$, and its estimate is accurate, so $|\hat{g}(\vec{s})| \leq |\hat{f}(\vec{s})| + \theta^2/4 \leq \frac{3}{2}|\hat{f}(\vec{s})|$. 
	By Parseval's identity, it follows that $W(g) \leq 9/4$.
	Since a coordinate of support size at most $k$ contributes to at most $k$ restricted influences, $\sum_{i} \mathrm{Inf}_i^{k}[g] \leq 9k/4$,
	so at most $|I| \leq (9k/4)/(\epsilon'/k) = 63k^2/(8\epsilon)$ features pass the threshold $\epsilon'/k$.
	The exhaustive search therefore examines at most $(63ek/(8\epsilon))^k$ subsets, each evaluated from the $L_0(g) = O(c^k/\epsilon^2)$ coefficients of $g$.
\end{proof}


\section{Experimental Evaluation}
\label{sec:experiments}

Our experiments address three questions.
(Q1)~Does the Spectral Explainer (\textsc{SE}, Algorithm~\ref{alg:se}) return explanations within $\epsilon$ of the optimum, with a failure rate at most $\delta$, as guaranteed by Theorem~\ref{thm:explainability}?
(Q2)~Does its query complexity scale with $d$, $k$ and $c$ as predicted by the analysis?
(Q3)~How does it compare to the Greedy Explanation baseline (\textsc{GE}) inspired by~\cite{Bassan.ICLR.2026} on real classifiers, and are the returned explanations meaningful?
Both algorithms are implemented in \texttt{Python 3.14} with \texttt{PyTorch}, and were run on a workstation with an Intel Xeon Gold 6252 CPU, 384\,GB of RAM, and two NVIDIA Quadro RTX 8000 GPUs. The detailed protocol and complete results are in Appendix~\ref{app:experiments}.

\paragraph{Explainers.}
\textsc{SE} follows Algorithm~\ref{alg:se}: given $(\epsilon, \delta, k)$, it sets $\epsilon' = 2\epsilon/7$ and $\theta = \epsilon'/c^{k/2}$, runs IKM with confidence $\delta_0$ and $b$ blocks as in Algorithm~\ref{alg:ikm}, and prunes features at $\epsilon'/k$. The only deviation from the analysis is the block size $n = \lceil \kappa \cdot 64/\theta^4 \rceil$. While Lemma~\ref{lem:estimation} requires $\kappa = 1$, which is conservative, calibration on hidden parities showed that $\kappa = 0.003$ was the smallest value with no failures. We use $\kappa = 0.01$, that is, 100 times fewer samples than prescribed. Our experiments thus test a stronger claim than Theorem~\ref{thm:explainability}: if the failure rate stays within $\delta$ under this reduced budget, the guarantee holds \emph{a fortiori} for $\kappa = 1$, which only makes estimates more accurate. Three implementation choices leave the analysis unchanged (Appendix~\ref{app:parameters}): IKM estimates all prefixes of a given depth on a shared batch (Remark~\ref{app:rem:shared}), and discards coordinates of degree greater than $k$, which are never used afterwards; moreover, since $\mathsf{err}_{f,\mathcal{D}}$ is non-increasing under inclusion, \textsc{SE} completes its explanation up to $k$ features, by decreasing order of estimated influence, whenever fewer than $k$ features pass the pruning. The baseline \textsc{GE} performs forward selection: it starts from $K = \emptyset$ and, at each step, adds the feature $i$ that minimizes $\mathsf{err}_{f,\mathcal{D}}(K \cup \{i\})$, estimated on $5{,}000$ fresh instances per candidate, until $|K| = k$ or no feature further reduces the error.

\paragraph{Evaluation.}
Each explanation $K$ is scored by its exact error $\mathsf{err}_{f,\mathcal{D}}(K) = (1 - W(f_K))/2$, computed by enumerating $\vec{\Omega}$,\footnote{When $|\vec{\Omega}| > 2^{22}$, $\mathsf{err}_{f,\mathcal{D}}(K)$ is estimated on $10^6$ instances.} and compared to $\opt = \min_{|K'| \leq k} \mathsf{err}_{f,\mathcal{D}}(K')$, computed in closed form or by brute force. A run is a \emph{success} if $\mathsf{err}_{f,\mathcal{D}}(K) \leq \opt + \epsilon$. Unless otherwise specified, $\epsilon = 0.35$ and $\delta = 0.1$. Each synthetic setting is repeated $10$ times ($100$ times for the $\delta$-sweep), and each real-world setting $6$ times for every $k \in \{1, \dots, 5\}$.

\paragraph{Synthetic targets.}
All targets are defined on $\{0, \dots, c-1\}^d$ under the uniform distribution, and built from the hidden parity $\chi_S(\vec{x}) = \prod_{i \in S} (-1)^{x_i}$. For even $c$, the whole energy of $\chi_S$ lies on coordinates with support $S$, so that $\mathsf{err}_{f,\mathcal{D}}(K) = 0$ if $S \subseteq K$, and $1/2$ otherwise: no proper subset of $S$ explains $\chi_S$ better than $\emptyset$, and forward selection receives no signal. We consider three targets:
(i)~the hidden parity $\chi_S$ itself, with $|S| = k$;
(ii)~a \emph{noisy parity}, where the labels of $\chi_S$ ($|S| = 3$) are flipped on a fixed pseudo-random subset of density $\eta = 0.05$, so that $\opt \approx 2\eta(1 - \eta) \approx 0.093$, attained by $K = S$;
(iii)~the \emph{Synergy Trap}, defined by $f(\vec{x}) = 1$ if $x_{i^*} = 1$ and $f(\vec{x}) = \chi_S(\vec{x})$ otherwise, with $i^* \notin S$ and $|S| = k - 1$.
In the Synergy Trap, $\mathsf{err}_{f,\mathcal{D}}(\emptyset) = 3/8$ and $\mathsf{err}_{f,\mathcal{D}}(\{i^*\}) = 1/4$, while adding any proper subset of $S$ to $\{i^*\}$ leaves the error unchanged: the distractor $i^*$ is the only feature that helps on its own, yet only $\{i^*\} \cup S$ reaches $\mathsf{err}_{f,\mathcal{D}}(K) = 0$.

\paragraph{Real-world benchmarks.}
We used nine binary classification tasks from \texttt{OpenML} with $d$ ranging from $8$ to $20$, which allows the exact computation of $\opt$. Instances with a missing label were removed, as were attributes with any missing value, and the remaining attributes were binarized ($c = 2$) using a median split for numerical variables and a most-frequent-value test for categorical ones. For each dataset, $f$ is a multi-layer perceptron with one hidden layer of 100 ReLU units, trained on $80\%$ of the data, and $\mathcal{D}$ is the product of empirical marginals with Laplace smoothing.

\begin{figure*}[t]
	\centering
	\begin{subfigure}{0.32\textwidth}
		\centering
		\includegraphics[width=\textwidth]{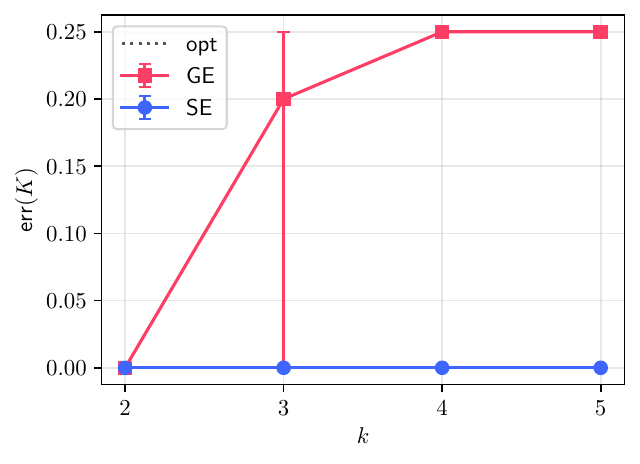}
		\caption{Synergy Trap}
		\label{fig:synthetic:trap}
	\end{subfigure}
	\hfill
	\begin{subfigure}{0.32\textwidth}
		\centering
		\includegraphics[width=\textwidth]{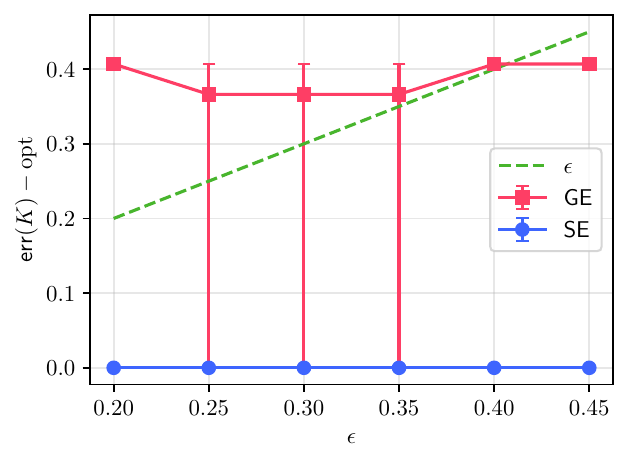}
		\caption{Noisy parity: gap}
		\label{fig:synthetic:eps}
	\end{subfigure}
	\hfill
	\begin{subfigure}{0.32\textwidth}
		\centering
		\includegraphics[width=\textwidth]{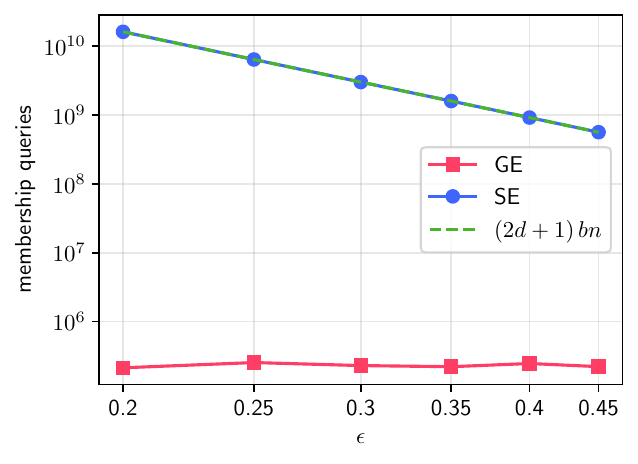}
		\caption{Noisy parity: queries}
		\label{fig:synthetic:queries}
	\end{subfigure}
	\caption{Synthetic targets ($d = 16$); bars show the range over runs.
		(a) $\mathsf{err}_{f,\mathcal{D}}(K)$ against $k$, with $\opt = 0$.
		(b) $\mathsf{err}_{f,\mathcal{D}}(K) - \opt$ against $\epsilon$, with the bound $\epsilon$ (dashed).
		(c) Membership queries against $\epsilon$, with the value $(2d+1)\,b\,n$ of the analysis (dashed).}
	\label{fig:synthetic}
\end{figure*}

\paragraph{Results on synthetic targets (Q1, Q2).}
\emph{Accuracy.}
The Synergy Trap ($d = 16$, Figure~\ref{fig:synthetic:trap}) clearly separates the two explainers: \textsc{SE} returns $\{i^*\} \cup S$ in all $40$ runs ($\mathsf{err}_{f,\mathcal{D}}(K) = 0$), while \textsc{GE} selects the distractor $i^*$ first and then stops, since no single feature of $S$ improves the error; it ends with $\mathsf{err}_{f,\mathcal{D}}(K) = 1/4$ in $28$ of the $30$ runs with $k \geq 3$. On the noisy parity ($d = 16$, $k = 3$, Figure~\ref{fig:synthetic:eps}), the gap $\mathsf{err}_{f,\mathcal{D}}(K) - \opt$ of \textsc{SE} is zero for all $\epsilon \in [0.20, 0.45]$, whereas \textsc{GE} misses the parity in most runs, with an average gap close to $0.4$. On hidden parities, \textsc{SE} reached $\opt$ in all $130$ runs of the cost sweeps below, against $7$ of the $120$ runs with $|S| \geq 2$ for \textsc{GE}.

\emph{Confidence.}
On the noisy parity, \textsc{SE} did not fail in any of $600$ runs for six values of $\delta$ ranging from $0.01$ to $0.5$ (Figure~\ref{fig:delta} in Appendix~\ref{app:synthetic}): the observed failure rate remains far below $\delta$, consistent with the conservative bound of Lemma~\ref{lem:estimation}.

\emph{Cost.}
In every run, the number of queries of \textsc{SE} is exactly $Q = (2d + 1)\, b\, n$ (Figure~\ref{fig:synthetic:queries}), as expected with shared batches. Table~\ref{tab:cost} decomposes the growth of $Q$ along three sweeps on hidden parities into the three factors of this product: $Q$ is linear in $d$ up to the logarithmic factor $b$, and grows as $n \propto \theta^{-4} = c^{2k}/\epsilon'^4$ in $k$ and $c$. The search itself remains cheap: IKM performs between $18$ and $251$ estimations on average, several orders of magnitude below its bound $8dc/\theta^2$, because the pruning at $\theta^2/2$ discards almost every prefix of a parity (Figure~\ref{fig:cost} in Appendix~\ref{app:synthetic}).

\begin{table}[t]
	\caption{Growth of the number of queries $Q = (2d+1)\,b\,n$ of \textsc{SE} along three sweeps on hidden parities ($\epsilon = 0.35$, $\delta = 0.1$): the observed factor is the product of the three predicted ones.}
	\label{tab:cost}
	\centering
	\small
	\begin{tabular}{lcccc}
		\toprule
		Sweep & $2d + 1$ & $n \propto c^{2k}$ & $b$ & $Q$ (observed) \\
		\midrule
		$d$: $8 \to 128$ ($k = 3$, $c = 2$) & $\times 15.1$ & $\times 1$ & $\times 1.20$ & $\times 18.1$ \\
		$k$: $k \to k + 1$ ($d = 16$, $c = 2$) & $\times 1$ & $\times 4$ & $\times 1.04$ to $\times 1.06$ & $\times 4.2$ \\
		$c$: $2 \to 6$ ($d = 16$, $k = 2$) & $\times 1$ & $\times 81$ & $\times 1.24$ & $\times 101$ \\
		\bottomrule
	\end{tabular}
\end{table}

\begin{table}[t]
	\caption{Real-world benchmarks ($\epsilon = 0.35$, $\delta = 0.1$).
		``$\opt$ reached'': runs with $\mathsf{err}_{f,\mathcal{D}}(K) = \opt$, over $k \in \{1, \dots, 5\}$ and $6$ runs.
		\textsc{SE} returned the same explanation in all runs, hence a single value. Time: one run of \textsc{SE} at $k = 5$.
		Acc.: test accuracy of $f$. All nine datasets are reported in Appendix~\ref{app:experiments}.}
	\label{tab:results}
	\begin{center}
	\setlength{\tabcolsep}{5pt}
	\begin{tabular}{lrrcccccr}
		\toprule
		& & & \multicolumn{2}{c}{$\opt$ reached} & \multicolumn{3}{c}{$\mathsf{err}_{f,\mathcal{D}}(K)$ at $k = 5$} & \\
		\cmidrule(lr){4-5} \cmidrule(lr){6-8}
		Dataset & $d$ & Acc. & \textsc{SE} & \textsc{GE} & \textsc{SE} & \textsc{GE} & $\opt$ & Time (s) \\
		\midrule
		Breast Cancer (WI) & 8  & 0.98 & 30/30 & 25/30 & 0.080 & $0.080 \pm 0.000$ & 0.080 & 244  \\
		Tic-Tac-Toe        & 9  & 0.94 & 30/30 & 21/30 & 0.249 & $0.254 \pm 0.007$ & 0.249 & 397  \\
		COMPAS             & 13 & 0.62 & 30/30 & 26/30 & 0.110 & $0.112 \pm 0.003$ & 0.110 & 949  \\
		Mushroom           & 20 & 1.00 & 24/30 & 17/30 & 0.183 & $0.185 \pm 0.003$ & 0.183 & 1998 \\
		Credit-G           & 20 & 0.70 & 24/30 & 21/30 & 0.217 & $0.221 \pm 0.004$ & 0.217 & 2464 \\
		\bottomrule
	\end{tabular}
	\end{center}
\end{table}

\paragraph{Results on real-world benchmarks (Q3).}
Table~\ref{tab:results} reports five of the nine datasets, as the remaining four yield similar conclusions. \textsc{SE} achieves the exact optimum in $138$ out of $150$ runs, compared to $110$ for \textsc{GE} ($246$ and $211$ out of $270$ across all datasets).
Every run of both explainers satisfies $\mathsf{err}_{f,\mathcal{D}}(K) \leq \opt + \epsilon$ by a wide margin: the largest gap is $0.004$ for \textsc{SE} and $0.018$ for \textsc{GE}. The few misses by \textsc{SE} are near-ties between explanations that the estimated spectrum does not resolve. \textsc{GE} misses more frequently, notably on Tic-Tac-Toe, where the label is a logical function with strong feature interactions. \textsc{SE} is also more stable: it returned the same explanation in all runs for each of the $45$ (dataset, $k$) pairs, compared to only $24$ for \textsc{GE}. This accuracy comes at a cost: since $n$ grows as $\theta^{-4}$, a run of \textsc{SE} at $k = 5$ takes between $4$ and $41$ minutes on these datasets, while \textsc{GE} completes in a fraction of a second.

\begin{table}[t]
	\caption{Explanations returned by \textsc{SE} at $k = 3$, all optimal and identical across runs.
	Each feature is denoted by the binary test that defines it; the explanation partitions $\vec{\Omega}$ into the $2^3 = 8$ regions defined by the outcomes of its three tests. The other four datasets are in Appendix~\ref{app:datasets}.}
	\label{tab:explanations:main}
	\begin{center}
	\small\setlength{\tabcolsep}{4pt}
	\begin{tabular}{lp{0.645\textwidth}c}
		\toprule
		Dataset & Explanation $K$ returned by \textsc{SE} ($k = 3$) & $\mathsf{err}_{f,\mathcal{D}}(K)$ \\
		\midrule
		Breast Cancer (WI) & clump thickness $> 4$, cell shape uniformity $> 1$, bland chromatin $> 3$ & 0.139 \\
		Tic-Tac-Toe        & top-left $=$ x, middle-middle $=$ x, bottom-left $=$ x & 0.326 \\
		COMPAS             & age $> 31$, priors count $> 2$, age category $=$ less than 25 & 0.138 \\
		Mushroom           & odor $=$ none, gill size $=$ broad, stalk surface above ring $=$ smooth & 0.235 \\
		Credit-G           & checking status $=$ no checking, duration $> 18$, savings $< 100$ & 0.247 \\
		\bottomrule
	\end{tabular}
	\end{center}
\end{table}

\paragraph{Explanations.}
Table~\ref{tab:explanations:main} illustrates the explanations returned by \textsc{SE} for $k = 3$.
Recall that a ``global'' explanation is a set of features, not a conjunction of literals: it states which tests suffice to determine the prediction of $f$ on average, whatever their outcomes. The resulting sets are directly interpretable. For example, in Tic-Tac-Toe, $f$ is best explained by the center and two corner squares, which lie on four and three winning lines respectively, more than any other square. This triple also reveals feature interactions: greedy selection always picks the center first, but then selects the wrong corner in three out of six runs at $k = 2$, and misses the optimal triple once at $k = 3$. In COMPAS, predictions depend mainly on age and the number of prior offenses; in Credit-G, the key features are checking account status, loan duration, and savings.


\section{Limitations and Perspectives}
\label{sec:conclusion}

We introduced a spectral framework for global explainability as a principled alternative to greedy selection. Using an integer extension of the KM algorithm, we established PAC-FPT guarantees for learning global explanations in discrete feature spaces. Experiments confirm these guarantees: \textsc{SE} recovers synergistic interactions missed by forward selection, reaches the exact optimum on most real-world benchmarks, and consistently returns the same explanation across runs.

Our study also reveals two main limitations. First, although \textsc{SE} is fixed-parameter tractable, its sample size grows as $\theta^{-4} \propto c^{2k}/\epsilon^4$ (Theorem~\ref{thm:convergence}): a run takes tens of minutes for $d = 20$ and $k = 5$, and several hours for $d \approx 60$. In the latter regime, the energy of $f$ is spread over many features, so that none passes pruning and the guarantee holds trivially, as $\epsilon$ exceeds the gain of any explanation over $\emptyset$. Still, our bounds leave room for improvement, since $300$ times fewer samples sufficed in calibration. Future work will explore tighter, adaptive spectral estimates, solver-inspired pruning, and faster prefix-tree traversal heuristics.

Second, our framework assumes a product distribution $\mathcal{D}$ over a discrete space $\vec{\Omega}$.
While this assumption is reasonable for many tabular tasks, it does not capture the complex dependencies present in real-world data.
Extending global explainability to distributions factorized by generative models---such as Bayesian networks
or probabilistic circuits---is a crucial direction for future research.
In this regard, Heidari and Khardon~\cite{Heidari.COLT.2025} recently extended the KM algorithm to Bayesian network distributions
over Boolean domains to demonstrate that DNF formulas are agnostically PAC learnable.
However, learning global probabilistic explanations in this broader context remains an open challenge.
Generalizing such spectral techniques to factorized distributions over integer-valued domains could yield explanations
that are more closely aligned with the true joint structure of the data.


\paragraph{Acknowledgments.}

We thank the anonymous reviewers for their constructive feedback, which helped enhance the clarity and presentation of this work. This research received support from the AI Chair EXPEKCTATION (ANR-19-CHIA-0005-01) and the JCJC project FORM-XAI (ANR-25-CE23-0146), both funded by the French National Research Agency. Additional support was provided by TAILOR, a project funded through the EU Horizon 2020 research and innovation programme under GA No 952215.


\bibliographystyle{plain}
\bibliography{neurips2026}


\newpage
\appendix


\section{Technical Appendix: Proofs and Algorithmic Details}
\label{app:theory}

This section provides the proofs of the main results stated in the paper, along with additional technical lemmas and details of the algorithms.


\subsection{Fourier Bases}

An orthonormal basis for $\mathcal{L}^2(\Omega, \mathcal D)$ can be constructed using the Stieltjes procedure (Algorithm~\ref{alg:stieltjes}),
which employs a three-term recurrence to generate orthonormal polynomials, where $\phi_j$ has degree $j$.
For ordinal features, this hierarchy is semantically rich: $\phi_0 = 1$ represents the constant component (the mean),
$\phi_1$ captures linear trends, and $\phi_2$ captures quadratic curvature.
For categorical features, the values $0, \ldots, c-1$ simply serve as distinct nodes, and the procedure still yields a valid basis.

\begin{algorithm}[h]
	\caption{Stieltjes Procedure}
	\label{alg:stieltjes}
	\DontPrintSemicolon
	\KwIn{Distribution $\mathcal D$ over $\Omega = \{0, 1, \ldots, c-1\}$}
	\KwOut{Orthonormal basis functions $\{\phi_0, \phi_1, \ldots, \phi_{c-1}\}$}
	\rule{\textwidth}{0.4pt}

	Initialize $\phi_{-1}(x) \leftarrow 0$, $\phi_0(x) \leftarrow 1$, and $\alpha_0 \leftarrow 0$\;
	\For{$j = 0$ \KwTo $c-2$}{
		$\beta_j \leftarrow \sum_{a=0}^{c-1} \mathcal D(a) \cdot a \cdot \phi_j(a)^2$ \tcp*{$\beta_j = \langle x\phi_j, \phi_j \rangle_{\mathcal D}$}
		$u_{j+1}(x) \leftarrow (x - \beta_j)\phi_j(x) - \alpha_j\phi_{j-1}(x)$ \tcp*{Orthogonalization step}
		$\alpha_{j+1} \leftarrow \sqrt{\sum_{a=0}^{c-1} \mathcal D(a) \cdot u_{j+1}(a)^2}$ \tcp*{Normalization}
		$\phi_{j+1}(x) \leftarrow u_{j+1}(x) / \alpha_{j+1}$\;
	}
	\Return $\{\phi_0, \phi_1, \ldots, \phi_{c-1}\}$\;
\end{algorithm}

\begin{lemma}
	\label{app:lem:local_basis}
	For any distribution $\mathcal D$ with full support over an alphabet $\Omega$ of size $c$,
	an orthonormal basis $\{\phi_0, \ldots, \phi_{c-1}\}$ for $\mathcal L^2(\Omega, \mathcal D)$ with $\phi_0 = 1$ can be computed in $O(c^2)$ time
	using the Stieltjes procedure~\cite{Gautschi.Book.2004}.
\end{lemma}
\begin{proof}
	We show by induction on $j$ that $\phi_0, \ldots, \phi_j$ are orthonormal and that each $\phi_k$ is a polynomial of degree exactly $k$.
	The base case $j = 0$ holds since $\langle \phi_0, \phi_0 \rangle_{\mathcal D} = \sum_{a} \mathcal D(a) = 1$.
	For the inductive step, note first that $\phi_0, \ldots, \phi_{j}$ span all polynomials of degree at most $j$,
	so $\phi_j$ is orthogonal to every polynomial of degree less than $j$.

	\emph{Well-definedness.} By construction, $u_{j+1}$ is a polynomial of degree exactly $j+1 \le c-1$.
	Since $\mathcal D$ has full support, $\alpha_{j+1}^2 = \sum_a \mathcal D(a)\, u_{j+1}(a)^2 = 0$ would force $u_{j+1}$ to vanish on all $c$ points of $\Omega$,
	which is impossible for a nonzero polynomial of degree at most $c - 1$. Hence $\alpha_{j+1} > 0$, and $\phi_{j+1}$ has unit norm and degree $j+1$.

	\emph{Orthogonality.} For $k \le j$, using $\langle x\phi_j, \phi_k \rangle_{\mathcal D} = \langle \phi_j, x\phi_k \rangle_{\mathcal D}$, we get
	\begin{align*}
		\langle u_{j+1}, \phi_k \rangle_{\mathcal D}
		= \langle \phi_j, x\phi_k \rangle_{\mathcal D} - \beta_j \Ind[k = j] - \alpha_j \Ind[k = j-1].
	\end{align*}
	If $k < j-1$, the polynomial $x\phi_k$ has degree $k+1 < j$, so all three terms vanish.
	If $k = j$, the first term equals $\beta_j$ by definition.
	If $k = j-1$, the recurrence at step $j-1$ gives $x\phi_{j-1} = \alpha_j \phi_j + \beta_{j-1}\phi_{j-1} + \alpha_{j-1}\phi_{j-2}$,
	so the first term equals $\alpha_j$. In all cases, $\langle u_{j+1}, \phi_k \rangle_{\mathcal D} = 0$.
	Finally, $c$ orthonormal functions in the $c$-dimensional space $\mathcal L^2(\Omega, \mathcal D)$ form a basis.

	Regarding runtime, each function is represented by its $c$ values on $\Omega$, so each of the $c-1$ iterations takes $O(c)$ time,
	for a total of $O(c^2)$.
\end{proof}

For each feature $i \in [d]$, let $\{\phi_{i,0}, \ldots, \phi_{i,c_i-1}\}$ be the orthonormal basis of $\mathcal{L}^2(\Omega_i, \mathcal{D}_i)$
given by Lemma~\ref{app:lem:local_basis}.
By taking the tensor product of these local bases, we derive a global Fourier basis for the product space $\vec{\Omega} = \Omega_1 \times \cdots \times \Omega_d$.
Under the product distribution $\mathcal D$, this basis enables a unique representation of any function $f : \vec{\Omega} \to \mathbb{R}$ as a linear combination of basis functions.
Let $\mathcal{S} = \prod_{i=1}^d \{0, 1, \ldots, c_i - 1\}$ denote the set of multi-indices, referred to as \emph{coordinates}.
The \emph{global basis function} of a multi-index $\vec{s} \in \mathcal{S}$ is defined as $\phi_{\vec{s}}(\vec{x}) = \prod_{i=1}^d \phi_{i,s_i}(x_i)$.

\begin{lemma}
	\label{app:lem:global_basis}
	Let $\mathcal D$ be a product distribution with full support over $\vec{\Omega}$.
	The collection $\{\phi_{\vec{s}} : \vec{s} \in \mathcal S\}$ forms a complete orthonormal basis for $\mathcal L^2(\vec{\Omega}, \mathcal D)$.
\end{lemma}
\begin{proof}
	For any pair $\vec{s}, \vec{t} \in \mathcal S$, the product structure of $\mathcal D$ and the definition of $\phi_{\vec{s}}$ yield:
	\begin{align*}
		\langle \phi_{\vec{s}}, \phi_{\vec{t}} \rangle_{\mathcal D}
		= \prod_{i=1}^d \Expectation_{x_i \sim \mathcal D_i} [\phi_{i,s_i}(x_i) \phi_{i,t_i}(x_i)] = \prod_{i=1}^d \Ind[s_i = t_i] = \Ind[\vec{s} = \vec{t}],
	\end{align*}
	following from the orthonormality of local bases (Lemma~\ref{app:lem:local_basis}).
	Since $\mathcal D$ has full support, the dimension of $\mathcal L^2(\vec{\Omega}, \mathcal D)$ is $|\vec{\Omega}| = \prod_{i} c_i = |\mathcal S|$,
	so this orthonormal set is complete.
\end{proof}

\begin{corollary}
	\label{app:cor:spectral_identities}
	Let $f, g \in \mathcal L^2(\vec{\Omega}, \mathcal D)$. Then, $\langle f, g \rangle_{\mathcal D} = \sum_{\vec{s} \in \mathcal S} \hat{f}(\vec{s}) \hat{g}(\vec{s})$.
	Furthermore, $\|f\|_2^2 = L_2(f)^2$, which equals $1$ if $f$ takes values in $\{-1, +1\}$.
\end{corollary}
\begin{proof}
	The first equality (Plancherel's theorem) follows directly from the orthonormality of $\{\phi_{\vec{s}}\}_{\vec{s} \in \mathcal S}$ established in Lemma~\ref{app:lem:global_basis}.
	The second equality (Parseval's identity) is the case $f = g$.
	In particular, if $f$ is a classifier taking values in $\{-1, +1\}$, then $\|f\|_2^2 = \Expectation[f(\vec{x})^2] = 1$, so $L_2(f)^2 = 1$.
\end{proof}


\subsection{The Integer Kushilevitz--Mansour Algorithm}

To demonstrate both the efficiency and correctness of the IKM algorithm, we begin by presenting a concentration bound for the MoM estimator.
This estimator is particularly valuable for product distributions, where local basis functions can have large ranges, as it relies on variance instead of boundedness.
Although this result was already established in~\cite{Lugosi.FCM.2019}, we provide a self-contained proof for completeness.

\begin{proposition}
	\label{app:prop:MoM}
	Let $X$ be a random variable with mean $\mu$ and variance at most $\sigma^2 < \infty$, and let $t > 0$ and $\delta_0 \in (0,1)$.
	Given $b \geq 8 \ln(1/\delta_0)$ blocks of $n \geq 4\sigma^2/t^2$ i.i.d.\ copies of $X$ each,
	the MoM estimator $\hat{\mu} = \mathrm{median}(\bar{X}_1, \ldots, \bar{X}_b)$ satisfies $|\hat{\mu} - \mu| \leq t$ with probability at least $1 - \delta_0$,
	where $\bar{X}_j$ is the empirical mean of the $j$-th block.
\end{proposition}
\begin{proof}
	Each block mean satisfies $\Expectation[\bar{X}_j] = \mu$ and $\mathrm{Var}[\bar{X}_j] \leq \sigma^2/n$.
	By Chebyshev's inequality and the choice of $n$,
	\begin{align*}
		\Probability(|\bar{X}_j - \mu| > t) \leq \frac{\sigma^2}{n t^2} \leq \frac{1}{4}.
	\end{align*}
	Let $E_j = \Ind[|\bar{X}_j - \mu| > t]$ and $B = \sum_{j=1}^b E_j$. The variables $E_j$ are independent, since the blocks are disjoint, and $\Expectation[B] \leq b/4$.
	If $|\hat{\mu} - \mu| > t$, then at least half of the block means lie on the same side of $\mu$ at distance more than $t$, so $B \geq b/2$.
	By Hoeffding's inequality,
	\begin{align*}
		\Probability(|\hat{\mu} - \mu| > t) \leq \Probability\left(B - \Expectation[B] \geq \frac{b}{4}\right) \leq e^{-2b(1/4)^2} = e^{-b/8} \leq \delta_0,
	\end{align*}
	where the last inequality follows from the choice of $b$.
\end{proof}

Applying this result to a classifier $f$ taking values in $\{-1, +1\}$, we note that the random variables averaged by our estimators have variance at most $1$,
independently of the marginal probabilities of the distribution. This yields the following guarantees, which restate Lemma~\ref{lem:estimation}.

\begin{lemma}
	\label{app:lem:estimation}
	Let $f: \vec{\Omega} \to \{-1,+1\}$ be a classifier and $\mathcal{D}$ a product distribution with full support.
	For any prefix $\vec{\sigma}$ and coordinate $\vec{s}$, the random variables averaged by \textsc{Weight} and \textsc{Coefficient}
	have expectations $W_{\vec{\sigma}}$ and $\hat{f}(\vec{s})$, respectively, and variance at most $1$.
	Consequently, with $b \geq 8 \ln(1/\delta_0)$ blocks of $n \geq 64/\theta^4$ samples each, every call to \textsc{Weight} or \textsc{Coefficient}
	returns an estimate within $\theta^2/4$ of its target with probability at least $1 - \delta_0$.
\end{lemma}
\begin{proof}
	Once the expectations and the variance bound are established, the result follows from Proposition~\ref{app:prop:MoM} with $\sigma^2 = 1$ and $t = \theta^2/4$,
	since then $4\sigma^2/t^2 = 64/\theta^4$.

	\emph{Coefficient estimation.}
	The procedure \textsc{Coefficient} averages $X = f(\vec{z}) \phi_{\vec{s}}(\vec{z})$, where $\vec{z} \sim \mathcal{D}$.
	Its expectation is $\langle f, \phi_{\vec{s}} \rangle_{\mathcal{D}} = \hat{f}(\vec{s})$.
	Since $f^2 = 1$ and $\Expectation[\phi_{\vec{s}}^2] = 1$, we have $\Expectation[X^2] = 1$, hence $\mathrm{Var}[X] = 1 - \hat{f}(\vec{s})^2 \leq 1$.

	\emph{Weight estimation.}
	Let $p$ be the length of $\vec{\sigma}$. The procedure \textsc{Weight} averages
	$X = f(\vec{u}, \vec{v})\, \phi_{\vec{\sigma}}(\vec{u})\, f(\vec{u}', \vec{v})\, \phi_{\vec{\sigma}}(\vec{u}')$,
	where $\vec{u}, \vec{u}' \sim \mathcal{D}_{1:p}$ and $\vec{v} \sim \mathcal{D}_{p+1:d}$ are independent.
	Conditioning on $\vec{v}$ and using the independence of $\vec{u}$ and $\vec{u}'$,
	\begin{align*}
		\Expectation[X \mid \vec{v}] = h_{\vec{\sigma}}(\vec{v})^2,
		\quad \text{where} \quad
		h_{\vec{\sigma}}(\vec{v}) = \Expectation_{\vec{u}}[f(\vec{u}, \vec{v})\, \phi_{\vec{\sigma}}(\vec{u})].
	\end{align*}
	Expanding $f$ in the global basis and using the orthonormality of $\{\phi_{\vec{\sigma}'}\}$ over prefixes of length $p$,
	we get $h_{\vec{\sigma}}(\vec{v}) = \sum_{\vec{s} \in \mathcal{C}_{\vec{\sigma}}} \hat{f}(\vec{s})\, \phi_{\vec{s}_{p+1:d}}(\vec{v})$,
	where $\phi_{\vec{s}_{p+1:d}}(\vec{v}) = \prod_{i = p+1}^d \phi_{i,s_i}(v_i)$.
	These suffix basis functions are orthonormal under $\mathcal{D}_{p+1:d}$, so
	\begin{align*}
		\Expectation[X] = \Expectation_{\vec{v}}[h_{\vec{\sigma}}(\vec{v})^2] = \sum_{\vec{s} \in \mathcal{C}_{\vec{\sigma}}} \hat{f}(\vec{s})^2 = W_{\vec{\sigma}}.
	\end{align*}
	For the variance, using $f^2 = 1$ and the independence of $\vec{u}$ and $\vec{u}'$,
	\begin{align*}
		\Expectation[X^2] = \Expectation_{\vec{u}}[\phi_{\vec{\sigma}}(\vec{u})^2] \cdot \Expectation_{\vec{u}'}[\phi_{\vec{\sigma}}(\vec{u}')^2] = 1,
	\end{align*}
	hence $\mathrm{Var}[X] \leq 1$.
\end{proof}

The logical structure of the search procedure guarantees that the algorithm identifies all significant spectral energy while ensuring the resulting hypothesis remains sparse.
In what follows, we assume $c \geq 2$ (a feature with a single value is constant and can be discarded).

\begin{proposition}
	\label{app:prop:logic}
	Let $g$ be the hypothesis produced by the IKM algorithm with threshold $\theta \in (0,1]$ and confidence $\delta \in (0,1)$,
	that is, with $\delta_0 = \theta^2 \delta / (8dc)$, $b = \lceil 8 \ln(1/\delta_0) \rceil$, and $n = \lceil 64/\theta^4 \rceil$.
	Then, with probability at least $1 - \delta$, IKM makes at most $N = 8dc/\theta^2$ calls to \textsc{Weight} and \textsc{Coefficient},
	and the following properties hold:
	\begin{description}
		\item[\emph{Completeness:}] For every $\vec{s}$ such that $|\hat{f}(\vec{s})| \geq \theta$, we have $\vec{s} \in \mathrm{supp}(g)$ and $|\hat{g}(\vec{s}) - \hat{f}(\vec{s})| \leq \theta^2/4$.
		\item[\emph{Soundness:}] For every $\vec{s} \in \mathrm{supp}(g)$, we have $|\hat{f}(\vec{s})| \geq \theta/2$.
	\end{description}
\end{proposition}

\begin{proof}
	Call an estimate \emph{accurate} if it lies within $\theta^2/4$ of its target, and let $G$ be the event that the first $\min(N, T)$ calls are accurate,
	where $T$ is the (random) total number of calls.
	Each call uses fresh samples, so conditionally on the past, it is accurate with probability at least $1 - \delta_0$ by Lemma~\ref{app:lem:estimation}.
	By a union bound over at most $N$ calls, $\Probability(G) \geq 1 - N\delta_0 = 1 - \delta$.
	We now show that, on $G$, the algorithm makes at most $N$ calls, so that all its estimates are accurate, and that both properties hold.

	\emph{Counting.}
	On $G$, consider the calls made so far. A prefix $\vec{\sigma}$ is retained only if $\hat{W}_{\vec{\sigma}} \geq \theta^2/2$, which implies $W_{\vec{\sigma}} \geq \theta^2/4$.
	Prefix cones of a given length $p$ partition $\mathcal{S}$, so by Parseval's identity, their weights sum to $\|f\|_2^2 = 1$.
	Hence at most $4/\theta^2$ prefixes are retained at each depth.
	Each retained prefix of length $p < d$ spawns at most $c$ children, so \textsc{Weight} is called on at most $4c/\theta^2$ prefixes at each depth $p \geq 1$,
	plus once on $\vec{\lambda}$, and \textsc{Coefficient} is called on at most $4/\theta^2$ leaves. The total number of calls is thus at most
	\begin{align*}
		1 + \frac{4dc}{\theta^2} + \frac{4}{\theta^2} \leq \frac{8dc}{\theta^2} = N,
	\end{align*}
	where the inequality uses $dc \geq 2$ and $\theta \leq 1$.

	\emph{Completeness.}
	Suppose that $|\hat{f}(\vec{s})| \geq \theta$. For any prefix $\vec{\sigma}$ of $\vec{s}$, we have $\vec{s} \in \mathcal{C}_{\vec{\sigma}}$, so $W_{\vec{\sigma}} \geq \hat{f}(\vec{s})^2 \geq \theta^2$, and therefore
	\begin{align*}
		\hat W_{\vec{\sigma}} \geq W_{\vec{\sigma}} - \frac{\theta^2}{4} \geq \frac{3\theta^2}{4} > \frac{\theta^2}{2}.
	\end{align*}
	Hence the path to $\vec{s}$ is never pruned, and $\vec{s}$ is added to $g$ with $|\hat{g}(\vec{s}) - \hat{f}(\vec{s})| \leq \theta^2/4$.

	\emph{Soundness.}
	Suppose $\vec{s} \in \mathrm{supp}(g)$. At depth $d$, the cone of $\vec{s}$ is $\{\vec{s}\}$, so $W_{\vec{s}} = \hat{f}(\vec{s})^2$.
	Since $\vec{s}$ was retained, $\hat{W}_{\vec{s}} \geq \theta^2/2$, and therefore
	$\hat{f}(\vec{s})^2 \geq \theta^2/2 - \theta^2/4 = \theta^2/4$, that is, $|\hat{f}(\vec{s})| \geq \theta/2$.
\end{proof}

\begin{theorem}
	\label{app:thm:convergence}
	Let $f: \vec{\Omega} \to \{-1,+1\}$ be a classifier and $\mathcal{D}$ a product distribution with full support and maximum alphabet size $c$.
	Assume that each membership query to $f$ and each draw from a marginal $\mathcal{D}_i$ take $O(1)$ time.
	Then, for any threshold $\theta \in (0,1]$ and confidence level $\delta \in (0,1)$, the IKM algorithm makes
	$O\big(\frac{dc}{\theta^6} \ln \frac{dc}{\theta \delta}\big)$ membership queries, runs in time $O\big(dc^2 + \frac{d^2 c}{\theta^6} \ln \frac{dc}{\theta \delta}\big)$,
	and returns a hypothesis $g$ such that, with probability at least $1 - \delta$:
	\begin{align*}
		L_{\infty}(f - g) \leq \theta \quad \text{and} \quad L_0(g) \leq \frac{4}{\theta^2}.
	\end{align*}
\end{theorem}
\begin{proof}
	We work on the event of Proposition~\ref{app:prop:logic}, which has probability at least $1 - \delta$.
	By soundness, every $\vec{s} \in \mathrm{supp}(g)$ satisfies $\hat{f}(\vec{s})^2 \geq \theta^2/4$, so by Parseval's identity,
	\begin{align*}
		1 = \sum_{\vec{s} \in \mathcal{S}} \hat{f}(\vec{s})^2 \geq \sum_{\vec{s} \in \mathrm{supp}(g)} \hat{f}(\vec{s})^2 \geq L_0(g) \cdot \frac{\theta^2}{4},
	\end{align*}
	which yields $L_0(g) \leq 4/\theta^2$.
	For $\vec{s} \in \mathrm{supp}(g)$, the estimate is accurate, so $|\hat{g}(\vec{s}) - \hat{f}(\vec{s})| \leq \theta^2/4 \leq \theta$.
	For $\vec{s} \notin \mathrm{supp}(g)$, completeness implies $|\hat{f}(\vec{s})| < \theta$, and $\hat{g}(\vec{s}) = 0$.
	Thus, $L_{\infty}(f - g) \leq \theta$.

	For the complexity, IKM makes at most $N = O(dc/\theta^2)$ calls, each using $bn = O(\theta^{-4} \ln(1/\delta_0))$ samples,
	with at most two membership queries per sample. Since $\ln(1/\delta_0) = \ln\frac{8dc}{\theta^2 \delta} = O\big(\ln \frac{dc}{\theta\delta}\big)$,
	the number of queries is $O\big(\frac{dc}{\theta^6} \ln \frac{dc}{\theta \delta}\big)$.
	Each sample also requires drawing $d$ coordinates and evaluating a basis function, which is a product of at most $d$ local terms, in $O(d)$ time.
	Adding the $O(\sum_i c_i^2) = O(dc^2)$ preprocessing time of the Stieltjes procedure (Lemma~\ref{app:lem:local_basis}),
	the total running time is
	\begin{align*}
		O\left(dc^2 + \frac{d^2 c}{\theta^6} \ln \frac{dc}{\theta \delta}\right).
	\end{align*}
\end{proof}

\begin{remark}[Shared batches]
	\label{app:rem:shared}
	IKM can process the search depth by depth, estimating at depth $p$ the weights of all the examined prefixes of length $p$ on a single batch of $b$ blocks of $n$ samples,
	and all the coefficients on a single final batch.
	The prefixes examined at depth $p$ are determined by the batches of the previous depths, which are independent of the batch of depth $p$.
	Hence, conditionally on the past, each estimate is still accurate with probability at least $1 - \delta_0$ by Lemma~\ref{app:lem:estimation},
	and the proof of Proposition~\ref{app:prop:logic} applies verbatim, since its union bound does not require the estimates to be independent.
	Each batch of depth $1 \leq p \leq d$ uses two membership queries per sample, and the final batch one, so that IKM makes exactly $(2d+1)\,b\,n = O\big(\frac{d}{\theta^4} \ln \frac{dc}{\theta \delta}\big)$ queries,
	independently of the number of examined prefixes. This is the implementation used in Section~\ref{sec:experiments}.
\end{remark}

\subsection{The Spectral Explanation Algorithm}

To derive the main result of this paper, we need several preparatory lemmas.
The first lemma establishes a structural identity, revealing that the global relevance of any feature subset is precisely quantified by its Fourier energy.

\begin{lemma}
	\label{app:lem:characterization}
	For any classifier $f: \vec{\Omega} \to \{-1, +1\}$, any product distribution $\mathcal{D}$ with full support on $\vec{\Omega}$, and any feature subset $K$, the global relevance error of $K$ satisfies:
	\begin{align*}
		\mathsf{err}_{f, \mathcal{D}}(K) = \frac{1}{2}\left(1 - W(f_K)\right).
	\end{align*}
\end{lemma}
\begin{proof}
	Let $\vec{x} \sim \mathcal{D}$ and $\vec{z} \sim \mathcal{D}(\cdot \mid \vec{z}_K = \vec{x}_K)$.
	Since $f$ takes values in $\{-1, +1\}$, we have $\Ind[f(\vec{x}) \neq f(\vec{z})] = (1 - f(\vec{x})f(\vec{z}))/2$, so
	\begin{align*}
		\mathsf{err}_{f, \mathcal{D}}(K) = \frac{1}{2}\left(1 - \Expectation[f(\vec{x})f(\vec{z})]\right).
	\end{align*}
	Conditioning on $\vec{x}$, we get $\Expectation[f(\vec{z}) \mid \vec{x}] = f_K(\vec{x})$ by definition of the projection. Hence, by Plancherel's theorem,
	\begin{align*}
		\Expectation[f(\vec{x})f(\vec{z})] = \Expectation_{\vec{x} \sim \mathcal{D}}[f(\vec{x}) f_K(\vec{x})] = \langle f, f_K \rangle_{\mathcal{D}} = \sum_{\vec{s} \in \mathcal{S}_K} \hat{f}(\vec{s})^2 = W(f_K),
	\end{align*}
	which yields the result.
\end{proof}

The next lemma ensures that the spectral energy of the hypothesis $g$ returned by the IKM algorithm
remains a reliable proxy for the energy of $f$ across all possible explanations of size $k$.
For a real-valued function $g$, the projection $g_K$ is defined as for classifiers, namely $g_K = \sum_{\vec{s} \in \mathcal{S}_K} \hat{g}(\vec{s}) \phi_{\vec{s}}$.

\begin{lemma}
	\label{app:lem:recovery}
	Let $f: \vec{\Omega} \to \{-1, +1\}$ be a classifier and $g: \vec{\Omega} \to \mathbb{R}$ such that $L_{\infty}(f - g) \leq \epsilon / c^{k/2}$ for some $\epsilon \in (0,1)$.
	Then, for any subset $K \subseteq [d]$ of size $|K| \leq k$, the difference between their projected energies satisfies $|W(f_K) - W(g_K)| \leq 3 \epsilon$.
\end{lemma}
\begin{proof}
	Since $|\mathcal{S}_K| = \prod_{i \in K} c_i \leq c^k$, Parseval's identity gives
	\begin{align*}
		\|f_K - g_K\|_2^2 = \sum_{\vec{s} \in \mathcal{S}_K} (\hat{f}(\vec{s}) - \hat{g}(\vec{s}))^2 \leq c^k \cdot \frac{\epsilon^2}{c^k} = \epsilon^2.
	\end{align*}
	Moreover, $\|f_K\|_2 \leq \|f\|_2 = 1$ and, by the triangle inequality, $\|g_K\|_2 \leq 1 + \epsilon$. Therefore,
	\begin{align*}
		|W(f_K) - W(g_K)|
		 & = \big|\|f_K\|_2 - \|g_K\|_2\big| \cdot \big(\|f_K\|_2 + \|g_K\|_2\big)            \\
		 & \leq \|f_K - g_K\|_2 \, (2 + \epsilon) \leq \epsilon(2 + \epsilon) \leq 3\epsilon,
	\end{align*}
	where the last inequality uses $\epsilon < 1$.
\end{proof}

The final lemma establishes that features with low restricted influence can be safely disregarded.

\begin{lemma}
	\label{app:lem:pruning}
	For a function $g: \vec{\Omega} \to \mathbb{R}$ and $\epsilon > 0$, let $I = \{i \in [d] : \mathrm{Inf}_i^{k}[g] \geq \epsilon/k\}$ be the set of influential features.
	Then for any $K \subseteq [d]$ with $|K| \leq k$, it holds that $W(g_K) \leq W(g_{K \cap I}) + \epsilon$.
\end{lemma}
\begin{proof}
	Since $\mathcal{S}_{K \cap I} \subseteq \mathcal{S}_K$, we have $W(g_K) - W(g_{K \cap I}) = W_{\mathcal{S}_K \setminus \mathcal{S}_{K \cap I}}(g)$.
	Any coordinate $\vec{s} \in \mathcal{S}_K \setminus \mathcal{S}_{K \cap I}$ satisfies $|\mathrm{supp}(\vec{s})| \leq |K| \leq k$
	and contains at least one index $i \in K \setminus I$ in its support, so $\vec{s} \in \mathcal{R}_i^k$ for such an $i$. Hence,
	\begin{align*}
		W_{\mathcal{S}_K \setminus \mathcal{S}_{K \cap I}}(g) \leq \sum_{i \in K \setminus I} \mathrm{Inf}_i^{k}[g] < |K \setminus I| \cdot \frac{\epsilon}{k} \leq \epsilon,
	\end{align*}
	where the strict inequality follows from the definition of $I$ (the bound holds trivially if $K \subseteq I$).
\end{proof}

Based on these lemmas, we are now in a position to present the main result.

\begin{theorem}
	\label{app:thm:explainability}
	There exists an algorithm such that, for any $d$-dimensional feature space $\vec{\Omega}$ with maximum alphabet size $c$,
	any binary classifier $f$ over $\vec{\Omega}$, and any product distribution $\mathcal{D}$ with full support over $\vec{\Omega}$,
	given any size constraint $k \leq d$, parameters $\epsilon, \delta \in (0,1)$, and membership query access to $f$,
	the algorithm returns a subset $K \subseteq [d]$ of size at most $k$ satisfying, with probability at least $1 - \delta$:
	\begin{align*}
		\mathsf{err}_{f, \mathcal{D}}(K) \leq \min_{K'\subseteq [d], |K'| \leq k} \mathsf{err}_{f, \mathcal{D}}(K') + \epsilon.
	\end{align*}
	The algorithm is fixed-parameter tractable in $k$, $c$, and $1/\epsilon$, with a runtime polynomial in $d$, $(ck/\epsilon)^k$, and $\ln(1/\delta)$.
\end{theorem}
\begin{proof}
	Let $K$ be the subset returned by Algorithm~\ref{alg:se}, with $\epsilon' = 2\epsilon/7$ and $\theta = \epsilon'/c^{k/2}$,
	and let $K^*$ maximize $W(f_{K^*})$ among subsets of size at most $k$.
	By Lemma~\ref{app:lem:characterization}, $K^*$ minimizes the global relevance error among such subsets, and
	\begin{align*}
		\mathsf{err}_{f, \mathcal{D}}(K) - \mathsf{err}_{f, \mathcal{D}}(K^*) = \frac{1}{2}\left(W(f_{K^*}) - W(f_K)\right),
	\end{align*}
	so it suffices to show that $W(f_{K^*}) - W(f_K) \leq 2\epsilon$.
	By Theorem~\ref{app:thm:convergence}, with probability at least $1 - \delta$, the IKM phase returns a hypothesis $g$ with $L_\infty(f - g) \leq \theta$
	and satisfying the properties of Proposition~\ref{app:prop:logic}. We assume this event in what follows, and decompose the energy gap as:
	\begin{align*}
		W(f_{K^*}) - W(f_K)
		 & = \underbrace{W(f_{K^*}) - W(g_{K^*})}_{\text{(A)}} + \underbrace{W(g_{K^*}) - W(g_{K^* \cap I})}_{\text{(B)}} \\
		 & + \underbrace{W(g_{K^* \cap I}) - W(g_K)}_{\text{(C)}} + \underbrace{W(g_K) - W(f_K)}_{\text{(D)}}.
	\end{align*}
	By Lemma~\ref{app:lem:recovery} with $\epsilon'$ in place of $\epsilon$, the recovery errors (A) and (D) are each bounded by $3\epsilon'$.
	By Lemma~\ref{app:lem:pruning} with the threshold $\epsilon'/k$, the pruning error (B) is bounded by $\epsilon'$.
	Since $K$ maximizes $W(g_{K'})$ over subsets $K' \subseteq I$ of size at most $k$, and $K^* \cap I$ is such a subset, the optimality gap (C) is non-positive.
	Summing these terms yields
	\begin{align*}
		W(f_{K^*}) - W(f_K) \leq 3\epsilon' + \epsilon' + 0 + 3\epsilon' = 7\epsilon' = 2\epsilon.
	\end{align*}

	For the runtime, we know by Theorem~\ref{app:thm:convergence} with $\theta = 2\epsilon/(7c^{k/2})$ that the IKM phase makes $O\big(d c^{3k+1} \epsilon^{-6} \ln \frac{dc^k}{\epsilon\delta}\big)$ queries
	and runs in time $O\big(dc^2 + d^2 c^{3k+1} \epsilon^{-6} \ln \frac{dc^k}{\epsilon\delta}\big)$.
	It returns at most $L_0(g) \leq 4/\theta^2 = 49 c^k/\epsilon^2$ coefficients, from which all restricted influences are computed in $O(d\, L_0(g))$ time.

	To bound the size of $I$, we use the event of Proposition~\ref{app:prop:logic}, assumed above.
	On this event, every $\vec{s} \in \mathrm{supp}(g)$ satisfies $|\hat{g}(\vec{s}) - \hat{f}(\vec{s})| \leq \theta^2/4$, since all estimates are accurate,
	and $|\hat{f}(\vec{s})| \geq \theta/2$, by soundness.
	Hence $|\hat{g}(\vec{s})| \leq |\hat{f}(\vec{s})| + \frac{\theta}{2}|\hat{f}(\vec{s})| \leq \frac{3}{2}|\hat{f}(\vec{s})|$, using $\theta \leq 1$.
	By Parseval's identity, $W(g) \leq \frac{9}{4} \sum_{\vec{s}} \hat{f}(\vec{s})^2 = \frac{9}{4}$.
	Each coordinate $\vec{s}$ with $|\mathrm{supp}(\vec{s})| \leq k$ contributes to at most $k$ restricted influences, so
	\begin{align*}
		\sum_{i = 1}^d \mathrm{Inf}_i^{k}[g] \leq k\, W(g) \leq \frac{9k}{4},
		\quad \text{and therefore} \quad
		|I| \leq \frac{9k/4}{\epsilon'/k} = \frac{63 k^2}{8\epsilon}.
	\end{align*}
	The exhaustive search examines all subsets of $I$ of size at most $k$. Their number is at most $(e|I|/k)^k \leq (63ek/(8\epsilon))^k$ when $|I| \geq k$,
	and at most $2^{|I|} \leq 2^k$ otherwise; in both cases, it is at most $(63ek/(8\epsilon))^k$.
	Before the search, we keep only the coordinates of $g$ whose support is included in $I$ and has at most $k$ elements, which takes $O(d\, L_0(g))$ time.
	Each energy $W(g_K)$ is then the sum of $\hat{g}(\vec{s})^2$ over the kept coordinates with support included in $K$, computed in $O(k\, L_0(g))$ time.
	The search thus takes $O\big(k\, L_0(g) \cdot (63ek/(8\epsilon))^k\big) = O\big(k c^k \epsilon^{-2} (63ek/(8\epsilon))^k\big)$ time.
	Adding the cost of the IKM phase and of the influence computation, the total runtime is polynomial in $d$, $(ck/\epsilon)^k$, and $\ln(1/\delta)$,
	which is fixed-parameter tractable in $k$, $c$, and $1/\epsilon$.
\end{proof}


\section{Experimental Appendix: Detailed Protocol and Extended Results}
\label{app:experiments}

This appendix details the protocol of Section~\ref{sec:experiments} and reports the complete results.
The code, the raw results of every run, and the commands that produced them are provided in the supplementary material.

\subsection{Parameters and Implementation of \textsc{SE}}
\label{app:parameters}

\paragraph{Parameters.}
\textsc{SE} takes as input $(\epsilon, \delta, k)$, the dimension $d$ and the alphabet size $c$, and derives all its parameters as in Algorithm~\ref{alg:se}:
\[
	\epsilon' = \frac{2\epsilon}{7}, \qquad
	\theta = \frac{\epsilon'}{c^{k/2}}, \qquad
	\delta_0 = \frac{\theta^2 \delta}{8dc}, \qquad
	b = \left\lceil 8 \ln \frac{1}{\delta_0} \right\rceil, \qquad
	n = \left\lceil \kappa \cdot \frac{64}{\theta^4} \right\rceil.
\]
IKM retains a prefix when its estimated weight is at least $\theta^2/2$, and performs at most $N = 8dc/\theta^2$ estimations;
the pruning step keeps the features $i$ with $\mathrm{Inf}_i^k[g] \geq \epsilon'/k$.
Every estimate is a median of $b$ means of $n$ samples.
The constant $\kappa$ is the only quantity that departs from the analysis, which requires $\kappa = 1$ (Lemma~\ref{app:lem:estimation}).
We calibrated it once, on hidden parities, as the smallest value with no failure over five seeds ($\kappa = 0.003$),
and used $\kappa = 0.01$ throughout, which leaves a margin of about $3$.
For instance, with $\epsilon = 0.35$, $\delta = 0.1$, $c = 2$ and $k = 5$, we get $\theta \approx 0.0177$, $n = 6{,}553{,}600$ and $b$ between $122$ and $130$ for $d \in [8, 20]$.

\paragraph{Implementation choices.}
Three choices depart from a literal reading of Algorithm~\ref{alg:se} without affecting its guarantee.
\begin{itemize}
	\item \emph{Shared batch per depth.} At depth $p$, the weights of all prefixes of length $p$ are estimated on the same $b$ blocks of $n$ samples, instead of one batch per prefix.
	      As shown in Remark~\ref{app:rem:shared}, the guarantee is unchanged, and a run makes exactly $(2d + 1)\, b\, n$ membership queries.
	\item \emph{Degree cap.} IKM discards the prefixes whose support already holds more than $k$ features.
	      Both the restricted influences $\mathrm{Inf}_i^k[g]$ and the energies $W(g_K)$ with $|K| \leq k$ only involve coordinates of degree at most $k$, so the discarded coordinates are never used.
	\item \emph{Completion.} Since $K \subseteq K'$ implies $W(f_K) \leq W(f_{K'})$, the error $\mathsf{err}_{f,\mathcal{D}}$ is non-increasing under inclusion.
	      When fewer than $k$ features survive the pruning, \textsc{SE} completes its explanation up to $k$ features by decreasing estimated influence, at no additional query.
	      Among all reported runs, this only happened on Bank Marketing (all $k$) and Breast Cancer ($k = 1$).
\end{itemize}

\paragraph{Machine cap.}
The implementation can cap the number of queries of a run at $r \cdot T$, 
where $r$ is the query throughput measured on the target and $T$ is a time limit (20 minutes by default); the cap lowers $n$, leaves $b$ unchanged, and records the effective constant $\kappa$.
Since the running time also grows with the number of prefixes estimated at each depth, this cap bounds the number of queries rather than the running time, which may exceed $T$.
The cap was never reached in the runs reported in this paper, which all use $\kappa = 0.01$.

\paragraph{Software.}
The code, available at \url{https://github.com/Lounesbo/neurips-2026}, is written in \texttt{Python 3.14}. The classifiers are trained with \texttt{scikit-learn} (\texttt{MLPClassifier} with default hyperparameters), and their forward pass is reimplemented in \texttt{PyTorch}, so that membership queries are answered on the GPU in batches of up to $2^{22}$ instances.

\paragraph{Hardware.}
All experiments ran on a workstation with an Intel Xeon Gold 6252 CPU, 384\,GB of RAM and two NVIDIA Quadro RTX 8000 GPUs (48\,GB each), one run per GPU at a time.

\subsection{Evaluation and Baseline}
\label{app:evaluation}

\paragraph{Exact evaluation.}
Every explanation $K$ is scored by $\mathsf{err}_{f,\mathcal{D}}(K) = (1 - W(f_K))/2$.
When $|\vec{\Omega}| \leq 2^{22}$, which holds for all real-world benchmarks and all synthetic targets with $d \leq 16$, 
the value $W(f_K)$ is computed exactly by enumerating $\vec{\Omega}$.
For each assignment $\vec{a}$ of the features in $K$, the projection $f_K$ takes the value $f_K(\vec{a}) = S_{\vec{a}} / \mathcal{D}_K(\vec{a})$, where $S_{\vec{a}} = \sum_{\vec{z}_K = \vec{a}} f(\vec{z})\, \mathcal{D}(\vec{z})$ and $\mathcal{D}_K(\vec{a}) = \prod_{i \in K} \mathcal{D}_i(a_i)$, so that
$W(f_K) = \Expectation[f_K^2] = \sum_{\vec{a}} S_{\vec{a}}^2 / \mathcal{D}_K(\vec{a})$.
The optimum $\opt$ is then computed by brute force over all subsets of size $k$ (which suffices, by monotonicity), or in closed form for hidden parities.
When $|\vec{\Omega}| > 2^{22}$ (the $d$-sweep with $d \geq 32$), $\mathsf{err}_{f,\mathcal{D}}(K)$ is estimated on $10^6$ pairs of instances that agree on $K$.

\paragraph{Greedy baseline.}
\textsc{GE} implements forward selection~\cite{Bassan.ICLR.2026}.
At each step, it draws $5{,}000$ fresh instances, re-estimates the error of the current set $K$, and estimates the error of $K \cup \{i\}$ for every candidate $i$ on an independent resample, so that the influence of one candidate cannot leak into the estimate of another.
It stops when $|K| = k$ or when no candidate decreases the estimated error.
Over all our experiments, a run of \textsc{GE} uses between $5 \cdot 10^4$ and $2 \cdot 10^6$ queries, against $4 \cdot 10^7$ to $4 \cdot 10^{10}$ for \textsc{SE}.

\paragraph{Runs and seeds.}
Each run of each explainer uses its own random generator, seeded by the series, the setting, and the run index, so that every run can be reproduced on its own.
Synthetic settings are repeated $10$ times ($100$ times for the $\delta$-sweep), and real-world settings $6$ times for each $k \in \{1, \dots, 5\}$.

\subsection{Synthetic Targets: Confidence and Cost}
\label{app:synthetic}

\paragraph{Targets.}
The hidden parity is $\chi_S(\vec{x}) = \prod_{i \in S} (-1)^{x_i}$, with $S$ drawn at random.
For even $c$ and uniform marginals, each factor has zero mean, so the whole energy of $\chi_S$ lies on coordinates with support $S$:
$\mathsf{err}_{f,\mathcal{D}}(K) = 0$ if $S \subseteq K$ and $1/2$ otherwise.
The noisy parity flips the label of $\chi_S$ ($d = 16$, $|S| = 3$) on a fixed pseudo-random subset of density $\eta = 0.05$, defined by hashing $\vec{x}$ so that $f$ remains a function;
its optimum, computed by brute force, is $\opt = 0.093 \approx 2\eta(1-\eta)$.
In the Synergy Trap ($d = 16$, $|S| = k - 1$), $\mathsf{err}_{f,\mathcal{D}}(\emptyset) = 3/8$ and $\mathsf{err}_{f,\mathcal{D}}(\{i^*\}) = 1/4$,
while adding a proper subset of $S$ leaves the error unchanged; only $\{i^*\} \cup S$ reaches $\mathsf{err}_{f,\mathcal{D}}(K) = 0$.

\begin{figure}[h]
	\centering
	\includegraphics[width=0.45\textwidth]{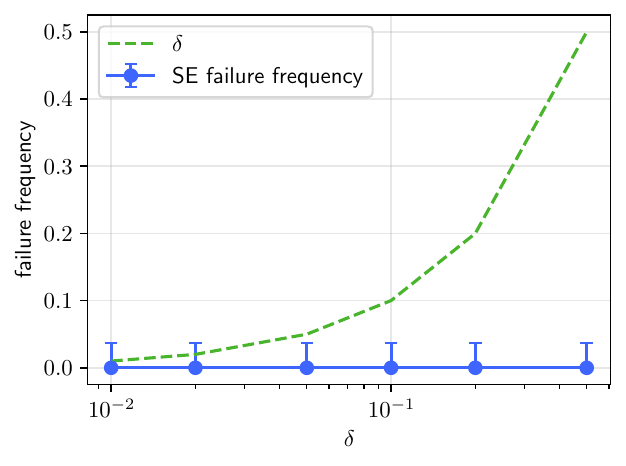}
	\caption{Failure frequency of \textsc{SE} against $\delta$ on the noisy parity ($d = 16$, $k = 3$, $\epsilon = 0.35$), with $95\%$ Wilson intervals over $100$ runs, and the bound $\delta$ (dashed).}
	\label{fig:delta}
\end{figure}

\paragraph{Confidence.}
Figure~\ref{fig:delta} reports the failure frequency of \textsc{SE} on the noisy parity, over $100$ runs for each $\delta \in \{0.01, 0.02, 0.05, 0.1, 0.2, 0.5\}$.
No run failed, and its gap to $\opt$ was zero in all $600$ runs; the $95\%$ Wilson intervals lie below $\delta$ for every $\delta \geq 0.05$.
By contrast, \textsc{GE} succeeded in only $2\%$ of these runs, with an average gap of $0.40$.

\begin{figure}[h]
	\centering
	\begin{subfigure}{\textwidth}
		\centering
		\includegraphics[width=0.8\textwidth]{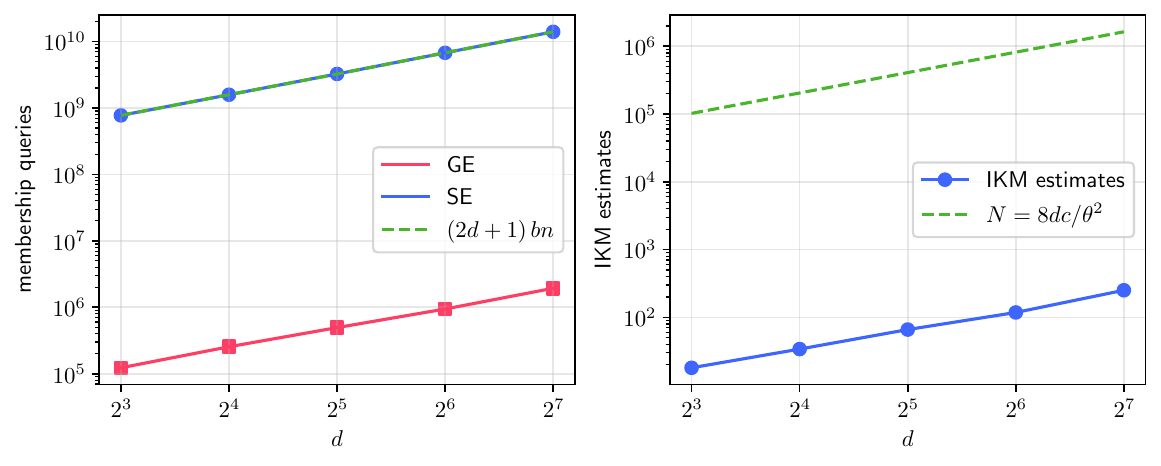}
		\caption{Dimension $d$ ($k = 3$, $c = 2$)}
	\end{subfigure}

	\begin{subfigure}{\textwidth}
		\centering
		\includegraphics[width=0.8\textwidth]{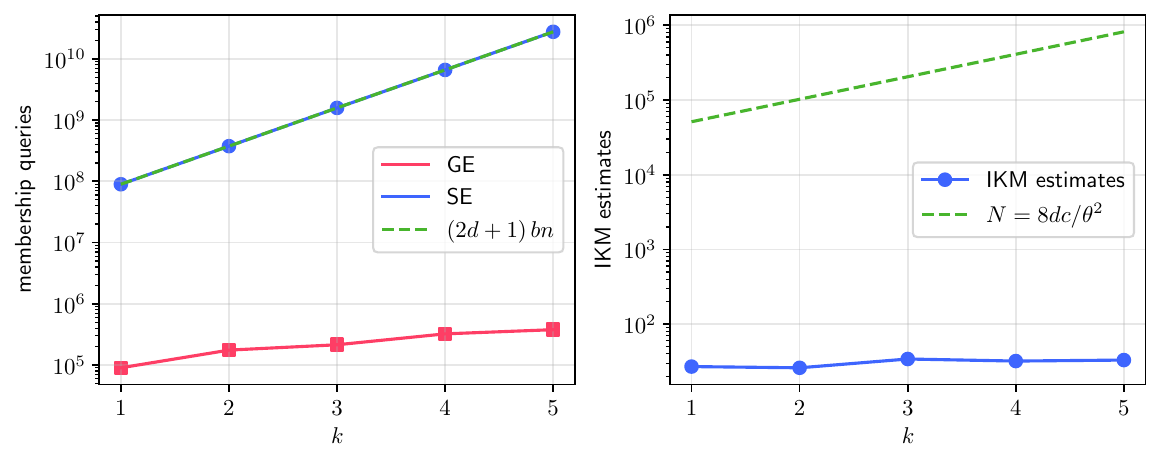}
		\caption{Explanation size $k$ ($d = 16$, $c = 2$)}
	\end{subfigure}

	\begin{subfigure}{\textwidth}
		\centering
		\includegraphics[width=0.8\textwidth]{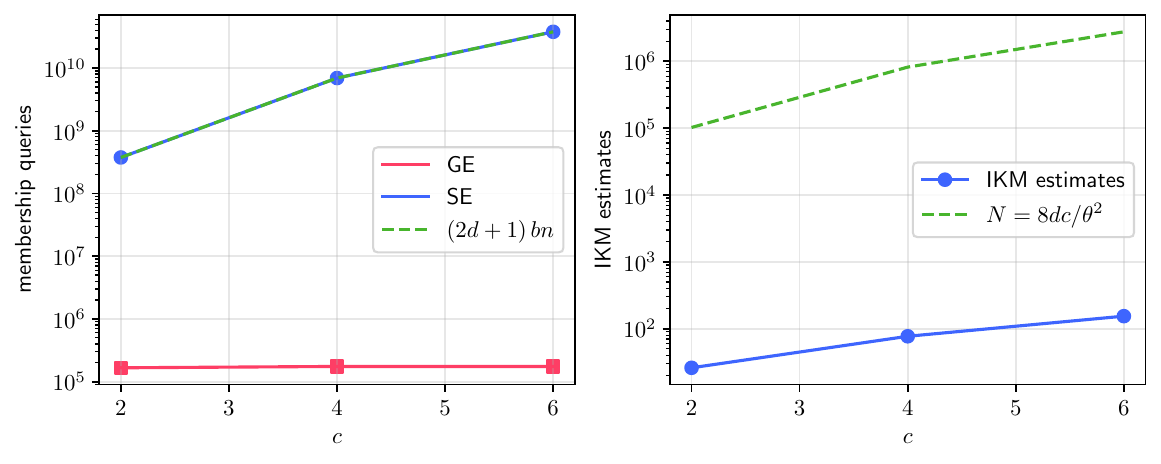}
		\caption{Alphabet size $c$ ($d = 16$, $k = 2$)}
	\end{subfigure}
	\caption{Cost of \textsc{SE} on hidden parities. Left: membership queries of \textsc{SE} and \textsc{GE}, with the value $(2d+1)\,b\,n$ of the analysis (dashed).
		Right: number of estimations performed by IKM, with the bound $N = 8dc/\theta^2$ (dashed). Both axes of the $d$-sweep are logarithmic.}
	\label{fig:cost}
\end{figure}

\paragraph{Cost.}
Figure~\ref{fig:cost} reports the cost of \textsc{SE} on hidden parities, where one parameter varies and the others are fixed ($\epsilon = 0.35$, $\delta = 0.1$; $k = 3$ and $c = 2$ for the $d$-sweep; $d = 16$ and $c = 2$ for the $k$-sweep; $d = 16$ and $k = 2$ for the $c$-sweep).
\textsc{SE} reached $\opt$ in all $130$ runs, while \textsc{GE} succeeded in only $7$ of the $120$ runs with $|S| \geq 2$, as expected on parities.
In every run, the number of queries is exactly $Q = (2d+1)\,b\,n$ (left panels), where $n \propto \theta^{-4} = c^{2k}/\epsilon'^4$ and $b = \lceil 8 \ln(8dc/(\theta^2\delta)) \rceil$ grows only logarithmically.
Table~\ref{tab:cost} (Section~\ref{sec:experiments}) decomposes the growth of $Q$ along each sweep into these three factors.
The number of weights and coefficients actually estimated by IKM (right panels) stays between $18$ and $251$ on average, several orders of magnitude below the bound $N = 8dc/\theta^2$,
because the pruning at $\theta^2/2$ discards almost every prefix of a parity.
The running time follows the same trends, from $1.6$\,s ($d = 8$) to $227$\,s ($d = 128$) at $k = 3$, and from $0.3$\,s ($k = 1$) to $93$\,s ($k = 5$) at $d = 16$.

\subsection{Real-World Benchmarks}
\label{app:datasets}

\paragraph{Preprocessing.}
The nine datasets are binary classification tasks from \texttt{OpenML}.
Instances with a missing label are removed, as are attributes with at least one missing value, and constant attributes.
Each remaining attribute is binarized ($c = 2$): a numerical attribute with more than two values is split at its median ($x > \text{median}$),
an attribute with two values is kept as is, and a categorical attribute with more values is reduced to a test on its most frequent value.
The data are split into $80\%$ for training and $20\%$ for testing, stratified by label.
$\mathcal{D}$ is the product of the empirical marginals of the binarized training and test data, with Laplace smoothing, which ensures full support.
Table~\ref{tab:datasets} summarizes the datasets.

\begin{table}[h]
	\caption{Real-world benchmarks. Raw: attributes of the \texttt{OpenML} dataset; Cat./Num.: categorical and numerical attributes kept;
		Dropped: attributes removed because of missing values or a constant value. The binary dimension is $d = \text{Cat.} + \text{Num.}$}
	\label{tab:datasets}
	\begin{center}
	\small
	\begin{tabular}{rlrrrrrcc}
		\toprule
		& & & & \multicolumn{2}{c}{Attributes} & & \multicolumn{2}{c}{Accuracy of $f$} \\
		\cmidrule(lr){5-6} \cmidrule(lr){8-9}
		ID & Dataset & Instances & Raw & Cat. & Num. & Dropped & Train & Test \\
		\midrule
		37     & Pima Indians Diabetes & 768    & 8  & 0  & 8  & 0 & 0.76 & 0.69 \\
		15     & Breast Cancer (WI)    & 699    & 9  & 0  & 8  & 1 & 0.96 & 0.98 \\
		50     & Tic-Tac-Toe           & 958    & 9  & 9  & 0  & 0 & 0.98 & 0.94 \\
		43,672 & Heart Disease         & 1,190  & 11 & 0  & 11 & 0 & 0.86 & 0.79 \\
		179    & Adult                 & 48,842 & 14 & 9  & 2  & 3 & 0.83 & 0.83 \\
		42,192 & COMPAS                & 5,278  & 13 & 6  & 7  & 0 & 0.68 & 0.62 \\
		1,461  & Bank Marketing        & 45,211 & 16 & 9  & 7  & 0 & 0.89 & 0.89 \\
		24     & Mushroom              & 8,124  & 22 & 20 & 0  & 2 & 1.00 & 1.00 \\
		31     & Credit-G              & 1,000  & 20 & 13 & 7  & 0 & 0.86 & 0.70 \\
		\bottomrule
	\end{tabular}
	\end{center}
\end{table}

\begin{figure}[h]
	\centering
	\begin{subfigure}{0.32\textwidth}\includegraphics[width=\textwidth]{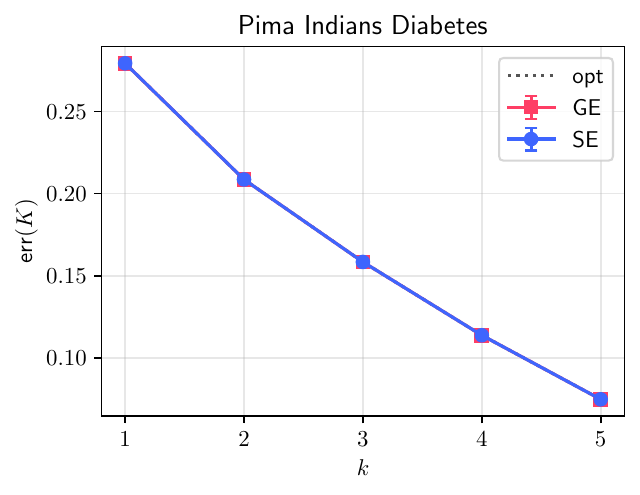}\end{subfigure}\hfill
	\begin{subfigure}{0.32\textwidth}\includegraphics[width=\textwidth]{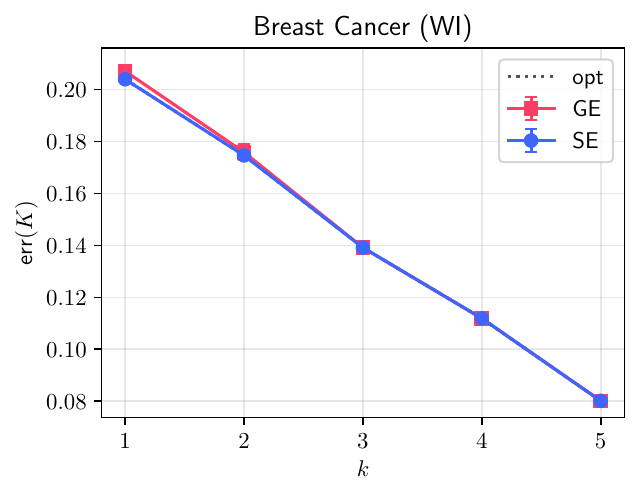}\end{subfigure}\hfill
	\begin{subfigure}{0.32\textwidth}\includegraphics[width=\textwidth]{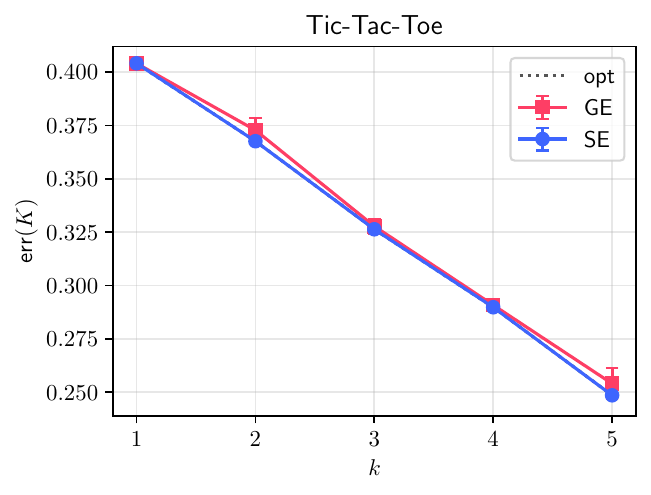}\end{subfigure}

	\begin{subfigure}{0.32\textwidth}\includegraphics[width=\textwidth]{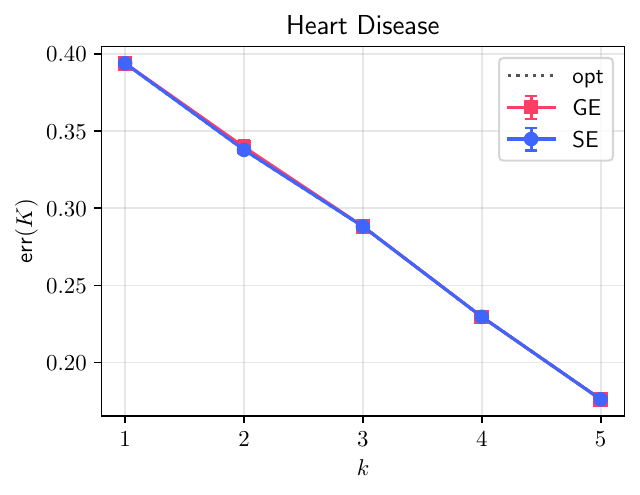}\end{subfigure}\hfill
	\begin{subfigure}{0.32\textwidth}\includegraphics[width=\textwidth]{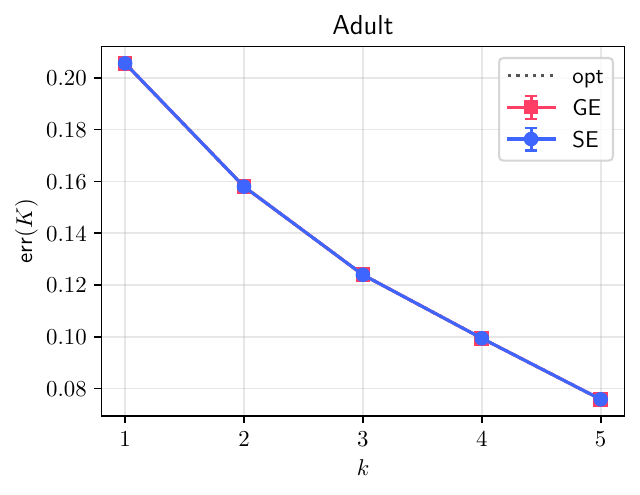}\end{subfigure}\hfill
	\begin{subfigure}{0.32\textwidth}\includegraphics[width=\textwidth]{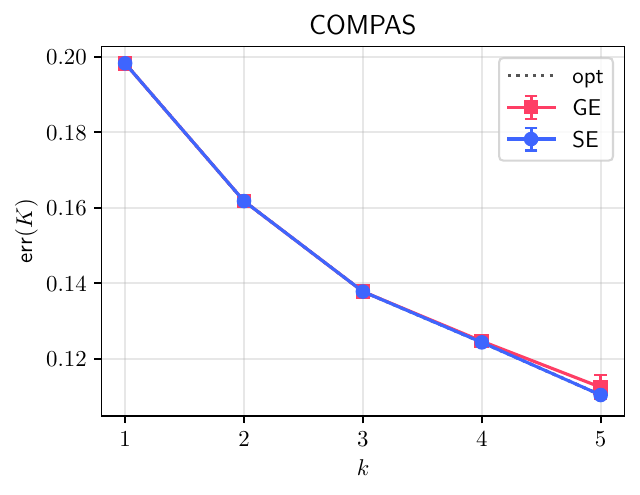}\end{subfigure}

	\begin{subfigure}{0.32\textwidth}\includegraphics[width=\textwidth]{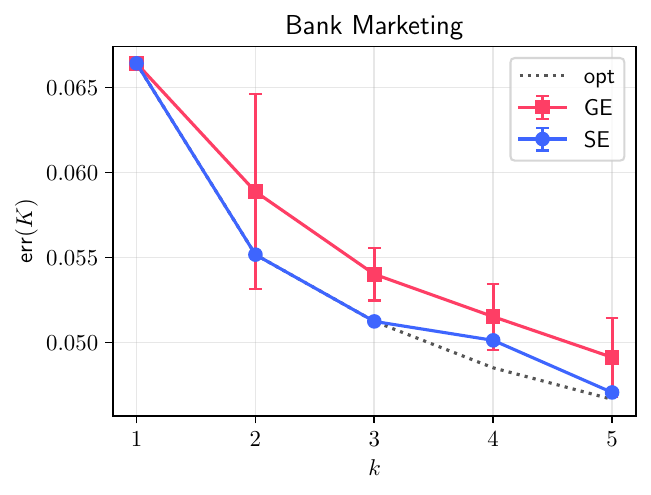}\end{subfigure}\hfill
	\begin{subfigure}{0.32\textwidth}\includegraphics[width=\textwidth]{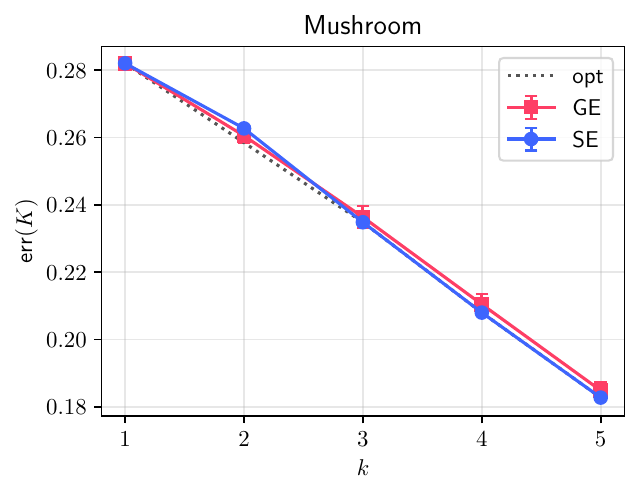}\end{subfigure}\hfill
	\begin{subfigure}{0.32\textwidth}\includegraphics[width=\textwidth]{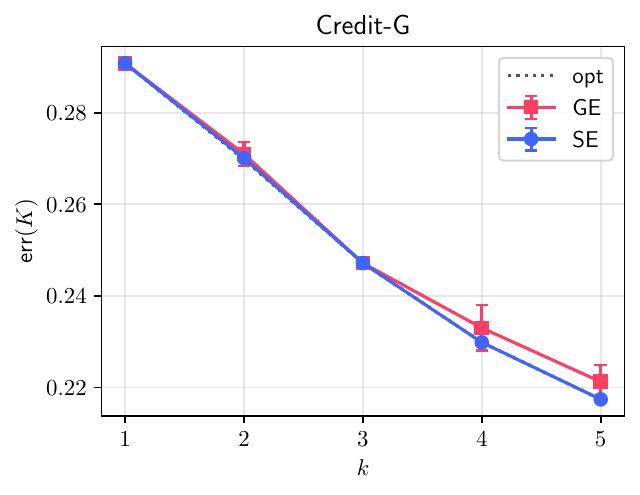}\end{subfigure}
	\caption{$\mathsf{err}_{f,\mathcal{D}}(K)$ against $k$ on the real-world benchmarks, with $\opt$ (dotted). Bars show the range over $6$ runs.}
	\label{fig:datasets}
\end{figure}

\paragraph{Complete results.}
Table~\ref{tab:results:all} extends Table~\ref{tab:results} to the nine datasets, and Figure~\ref{fig:datasets} shows $\mathsf{err}_{f,\mathcal{D}}(K)$ against $k$.
On Pima Indians Diabetes and Adult, both explainers reach $\opt$ in every run; on Heart Disease, \textsc{GE} misses it twice.
On Bank Marketing, the label is strongly imbalanced, so that most of the energy of $f$ lies on the constant coefficient:
at most two features pass the pruning threshold, and \textsc{SE} relies on the completion for every $k$.
It still reaches $\opt$ for $k \leq 3$, and misses it by $0.0016$ and $0.0004$ for $k = 4$ and $k = 5$, two near-ties like the other misses of \textsc{SE}. 
All runs of both explainers succeed.
For all datasets, the curves of \textsc{SE} coincide with $\opt$ for most $k$, while those of \textsc{GE} show a spread across runs.

\begin{table}[h]
	\caption{All real-world benchmarks ($\epsilon = 0.35$, $\delta = 0.1$); columns as in Table~\ref{tab:results}.}
	\label{tab:results:all}
	\begin{center}
	\small
	\setlength{\tabcolsep}{5pt}
	\begin{tabular}{lrrcccccr}
		\toprule
		& & & \multicolumn{2}{c}{$\opt$ reached} & \multicolumn{3}{c}{$\mathsf{err}_{f,\mathcal{D}}(K)$ at $k = 5$} & \\
		\cmidrule(lr){4-5} \cmidrule(lr){6-8}
		Dataset & $d$ & Acc. & \textsc{SE} & \textsc{GE} & \textsc{SE} & \textsc{GE} & $\opt$ & Time (s) \\
		\midrule
		Pima Indians Diabetes & 8  & 0.69 & 30/30 & 30/30 & 0.075 & $0.075 \pm 0.000$ & 0.075 & 248  \\
		Breast Cancer (WI)    & 8  & 0.98 & 30/30 & 25/30 & 0.080 & $0.080 \pm 0.000$ & 0.080 & 244  \\
		Tic-Tac-Toe           & 9  & 0.94 & 30/30 & 21/30 & 0.249 & $0.254 \pm 0.007$ & 0.249 & 397  \\
		Heart Disease         & 11 & 0.79 & 30/30 & 28/30 & 0.176 & $0.176 \pm 0.000$ & 0.176 & 722  \\
		Adult                 & 11 & 0.83 & 30/30 & 30/30 & 0.076 & $0.076 \pm 0.000$ & 0.076 & 580  \\
		COMPAS                & 13 & 0.62 & 30/30 & 26/30 & 0.110 & $0.112 \pm 0.003$ & 0.110 & 949  \\
		Bank Marketing        & 16 & 0.89 & 18/30 & 13/30 & 0.047 & $0.049 \pm 0.002$ & 0.047 & 843  \\
		Mushroom              & 20 & 1.00 & 24/30 & 17/30 & 0.183 & $0.185 \pm 0.003$ & 0.183 & 1998 \\
		Credit-G              & 20 & 0.70 & 24/30 & 21/30 & 0.217 & $0.221 \pm 0.004$ & 0.217 & 2464 \\
		\midrule
		Total                 &    &      & 246/270 & 211/270 & & & & \\
		\bottomrule
	\end{tabular}
	\end{center}
\end{table}

\paragraph{Explanations.}
Table~\ref{tab:explanations} lists the explanations returned by \textsc{SE} at $k = 3$; each of them is optimal, and identical across runs.
They are readable as such: for instance, on Adult, the prediction of the classifier is best explained by the education level, the marital status, and the capital gains of a person; on Tic-Tac-Toe, by the contents of the center and of two corners of the board.

\begin{table}[h]
	\caption{Explanations returned by \textsc{SE} at $k = 3$, all equal to the optimum.
	Each feature is denoted by the binary test that defines its binarized attribute.
	An explanation is a set of features, not a conjunction of literals:
	it partitions $\vec{\Omega}$ into the $2^3 = 8$ regions defined by the outcomes of its three tests.}	
	\label{tab:explanations}
	\begin{center}
	\small
	\begin{tabular}{lp{0.58\textwidth}c}
		\toprule
		Dataset & Explanation $K$ returned by \textsc{SE} ($k = 3$) & $\mathsf{err}_{f,\mathcal{D}}(K)$ \\
		\midrule
		Pima Indians Diabetes & plasma glucose $> 117$, body mass index $> 32$, age $> 29$ & 0.158 \\
		Breast Cancer (WI)    & clump thickness $> 4$, cell shape uniformity $> 1$, bland chromatin $> 3$ & 0.139 \\
		Tic-Tac-Toe           & top-left $=$ x, middle-middle $=$ x, bottom-left $=$ x & 0.326 \\
		Heart Disease         & sex $= 1$, chest pain type $\geq 4$, exercise angina $= 1$ & 0.288 \\
		Adult                 & education-num $> 10$, marital status $=$ married-civ-spouse, capital gain $= 0$ & 0.124 \\
		COMPAS                & age $> 31$, priors count $> 2$, age category $=$ less than 25 & 0.138 \\
		Bank Marketing        & housing loan $=$ yes, call duration $> 180$, previous outcome $=$ unknown & 0.051 \\
		Mushroom              & odor $=$ none, gill size $=$ broad, stalk surface above ring $=$ smooth & 0.235 \\
		Credit-G              & checking status $=$ no checking, duration $> 18$, savings $< 100$ & 0.247 \\
		\bottomrule
	\end{tabular}
	\end{center}
\end{table}

\clearpage
\section*{NeurIPS Paper Checklist}

\begin{enumerate}

\item {\bf Claims}
    \item[] Question: Do the main claims made in the abstract and introduction accurately reflect the paper's contributions and scope?
    \item[] Answer: \answerYes{} 
    \item[] Justification: The abstract and introduction clearly articulate our main contributions, methodology, and key results. 
    \item[] Guidelines:
    \begin{itemize}
        \item The answer \answerNA{} means that the abstract and introduction do not include the claims made in the paper.
        \item The abstract and/or introduction should clearly state the claims made, including the contributions made in the paper and important assumptions and limitations. A \answerNo{} or \answerNA{} answer to this question will not be perceived well by the reviewers. 
        \item The claims made should match theoretical and experimental results, and reflect how much the results can be expected to generalize to other settings. 
        \item It is fine to include aspirational goals as motivation as long as it is clear that these goals are not attained by the paper. 
    \end{itemize}

\item {\bf Limitations}
    \item[] Question: Does the paper discuss the limitations of the work performed by the authors?
    \item[] Answer: \answerYes{} 
    \item[] Justification: The last section of the paper includes a discussion of the limitations, addressing assumptions, scope, and potential impacts on performance.
    \item[] Guidelines:
    \begin{itemize}
        \item The answer \answerNA{} means that the paper has no limitation while the answer \answerNo{} means that the paper has limitations, but those are not discussed in the paper. 
        \item The authors are encouraged to create a separate ``Limitations'' section in their paper.
        \item The paper should point out any strong assumptions and how robust the results are to violations of these assumptions (e.g., independence assumptions, noiseless settings, model well-specification, asymptotic approximations only holding locally). The authors should reflect on how these assumptions might be violated in practice and what the implications would be.
        \item The authors should reflect on the scope of the claims made, e.g., if the approach was only tested on a few datasets or with a few runs. In general, empirical results often depend on implicit assumptions, which should be articulated.
        \item The authors should reflect on the factors that influence the performance of the approach. For example, a facial recognition algorithm may perform poorly when image resolution is low or images are taken in low lighting. Or a speech-to-text system might not be used reliably to provide closed captions for online lectures because it fails to handle technical jargon.
        \item The authors should discuss the computational efficiency of the proposed algorithms and how they scale with dataset size.
        \item If applicable, the authors should discuss possible limitations of their approach to address problems of privacy and fairness.
        \item While the authors might fear that complete honesty about limitations might be used by reviewers as grounds for rejection, a worse outcome might be that reviewers discover limitations that aren't acknowledged in the paper. The authors should use their best judgment and recognize that individual actions in favor of transparency play an important role in developing norms that preserve the integrity of the community. Reviewers will be specifically instructed to not penalize honesty concerning limitations.
    \end{itemize}

\item {\bf Theory assumptions and proofs}
    \item[] Question: For each theoretical result, does the paper provide the full set of assumptions and a complete (and correct) proof?
    \item[] Answer: \answerYes{} 
    \item[] Justification: The paper and the supplemental material provide all necessary assumptions and complete proofs for the theoretical results presented.
    \item[] Guidelines:
    \begin{itemize}
        \item The answer \answerNA{} means that the paper does not include theoretical results. 
        \item All the theorems, formulas, and proofs in the paper should be numbered and cross-referenced.
        \item All assumptions should be clearly stated or referenced in the statement of any theorems.
        \item The proofs can either appear in the main paper or the supplemental material, but if they appear in the supplemental material, the authors are encouraged to provide a short proof sketch to provide intuition. 
        \item Inversely, any informal proof provided in the core of the paper should be complemented by formal proofs provided in appendix or supplemental material.
        \item Theorems and Lemmas that the proof relies upon should be properly referenced. 
    \end{itemize}

    \item {\bf Experimental result reproducibility}
    \item[] Question: Does the paper fully disclose all the information needed to reproduce the main experimental results of the paper to the extent that it affects the main claims and/or conclusions of the paper (regardless of whether the code and data are provided or not)?
    \item[] Answer: \answerYes{} 
    \item[] Justification: The paper and the supplemental material provide all necessary details for reproducing the main experimental results, including dataset descriptions, model architectures, and training procedures.
    \item[] Guidelines:
    \begin{itemize}
        \item The answer \answerNA{} means that the paper does not include experiments.
        \item If the paper includes experiments, a \answerNo{} answer to this question will not be perceived well by the reviewers: Making the paper reproducible is important, regardless of whether the code and data are provided or not.
        \item If the contribution is a dataset and\slash or model, the authors should describe the steps taken to make their results reproducible or verifiable. 
        \item Depending on the contribution, reproducibility can be accomplished in various ways. For example, if the contribution is a novel architecture, describing the architecture fully might suffice, or if the contribution is a specific model and empirical evaluation, it may be necessary to either make it possible for others to replicate the model with the same dataset, or provide access to the model. In general. releasing code and data is often one good way to accomplish this, but reproducibility can also be provided via detailed instructions for how to replicate the results, access to a hosted model (e.g., in the case of a large language model), releasing of a model checkpoint, or other means that are appropriate to the research performed.
        \item While NeurIPS does not require releasing code, the conference does require all submissions to provide some reasonable avenue for reproducibility, which may depend on the nature of the contribution. For example
        \begin{enumerate}
            \item If the contribution is primarily a new algorithm, the paper should make it clear how to reproduce that algorithm.
            \item If the contribution is primarily a new model architecture, the paper should describe the architecture clearly and fully.
            \item If the contribution is a new model (e.g., a large language model), then there should either be a way to access this model for reproducing the results or a way to reproduce the model (e.g., with an open-source dataset or instructions for how to construct the dataset).
            \item We recognize that reproducibility may be tricky in some cases, in which case authors are welcome to describe the particular way they provide for reproducibility. In the case of closed-source models, it may be that access to the model is limited in some way (e.g., to registered users), but it should be possible for other researchers to have some path to reproducing or verifying the results.
        \end{enumerate}
    \end{itemize}

\item {\bf Open access to data and code}
    \item[] Question: Does the paper provide open access to the data and code, with sufficient instructions to faithfully reproduce the main experimental results, as described in supplemental material?
    \item[] Answer: \answerYes{} 
    \item[] Justification: All datasets are accessible through OpenML. The supplemental material provides access to the code, with scripts to faithfully reproduce the main experimental results.
    \item[] Guidelines:
    \begin{itemize}
        \item The answer \answerNA{} means that paper does not include experiments requiring code.
        \item Please see the NeurIPS code and data submission guidelines (\url{https://neurips.cc/public/guides/CodeSubmissionPolicy}) for more details.
        \item While we encourage the release of code and data, we understand that this might not be possible, so \answerNo{} is an acceptable answer. Papers cannot be rejected simply for not including code, unless this is central to the contribution (e.g., for a new open-source benchmark).
        \item The instructions should contain the exact command and environment needed to run to reproduce the results. See the NeurIPS code and data submission guidelines (\url{https://neurips.cc/public/guides/CodeSubmissionPolicy}) for more details.
        \item The authors should provide instructions on data access and preparation, including how to access the raw data, preprocessed data, intermediate data, and generated data, etc.
        \item The authors should provide scripts to reproduce all experimental results for the new proposed method and baselines. If only a subset of experiments are reproducible, they should state which ones are omitted from the script and why.
        \item At submission time, to preserve anonymity, the authors should release anonymized versions (if applicable).
        \item Providing as much information as possible in supplemental material (appended to the paper) is recommended, but including URLs to data and code is permitted.
    \end{itemize}

\item {\bf Experimental setting/details}
    \item[] Question: Does the paper specify all the training and test details (e.g., data splits, hyperparameters, how they were chosen, type of optimizer) necessary to understand the results?
    \item[] Answer: \answerYes{} 
    \item[] Justification: The paper and the supplemental material provide all necessary details for understanding the main experimental results.
    \item[] Guidelines:
    \begin{itemize}
        \item The answer \answerNA{} means that the paper does not include experiments.
        \item The experimental setting should be presented in the core of the paper to a level of detail that is necessary to appreciate the results and make sense of them.
        \item The full details can be provided either with the code, in appendix, or as supplemental material.
    \end{itemize}

\item {\bf Experiment statistical significance}
    \item[] Question: Does the paper report error bars suitably and correctly defined or other appropriate information about the statistical significance of the experiments?
    \item[] Answer: \answerYes{} 
    \item[] Justification: Yes, the paper and supplemental material report standard deviations correctly measured across multiple runs to rigorously assess the statistical significance of the results.
    \item[] Guidelines:
    \begin{itemize}
        \item The answer \answerNA{} means that the paper does not include experiments.
        \item The authors should answer \answerYes{} if the results are accompanied by error bars, confidence intervals, or statistical significance tests, at least for the experiments that support the main claims of the paper.
        \item The factors of variability that the error bars are capturing should be clearly stated (for example, train/test split, initialization, random drawing of some parameter, or overall run with given experimental conditions).
        \item The method for calculating the error bars should be explained (closed form formula, call to a library function, bootstrap, etc.)
        \item The assumptions made should be given (e.g., Normally distributed errors).
        \item It should be clear whether the error bar is the standard deviation or the standard error of the mean.
        \item It is OK to report 1-sigma error bars, but one should state it. The authors should preferably report a 2-sigma error bar than state that they have a 96\% CI, if the hypothesis of Normality of errors is not verified.
        \item For asymmetric distributions, the authors should be careful not to show in tables or figures symmetric error bars that would yield results that are out of range (e.g., negative error rates).
        \item If error bars are reported in tables or plots, the authors should explain in the text how they were calculated and reference the corresponding figures or tables in the text.
    \end{itemize}

\item {\bf Experiments compute resources}
    \item[] Question: For each experiment, does the paper provide sufficient information on the computer resources (type of compute workers, memory, time of execution) needed to reproduce the experiments?
    \item[] Answer: \answerYes{} 
    \item[] Justification: The paper provides detailed information on the computer resources used for the experiments, including the architecture, memory, and execution time.
    \item[] Guidelines:
    \begin{itemize}
        \item The answer \answerNA{} means that the paper does not include experiments.
        \item The paper should indicate the type of compute workers CPU or GPU, internal cluster, or cloud provider, including relevant memory and storage.
        \item The paper should provide the amount of compute required for each of the individual experimental runs as well as estimate the total compute. 
        \item The paper should disclose whether the full research project required more compute than the experiments reported in the paper (e.g., preliminary or failed experiments that didn't make it into the paper). 
    \end{itemize}
    
\item {\bf Code of ethics}
    \item[] Question: Does the research conducted in the paper conform, in every respect, with the NeurIPS Code of Ethics \url{https://neurips.cc/public/EthicsGuidelines}?
    \item[] Answer: \answerYes{} 
    \item[] Justification: The research conducted in the paper adheres to the NeurIPS Code of Ethics in all aspects.
    \item[] Guidelines:
    \begin{itemize}
        \item The answer \answerNA{} means that the authors have not reviewed the NeurIPS Code of Ethics.
        \item If the authors answer \answerNo, they should explain the special circumstances that require a deviation from the Code of Ethics.
        \item The authors should make sure to preserve anonymity (e.g., if there is a special consideration due to laws or regulations in their jurisdiction).
    \end{itemize}

\item {\bf Broader impacts}
    \item[] Question: Does the paper discuss both potential positive societal impacts and negative societal impacts of the work performed?
    \item[] Answer: \answerNA{} 
    \item[] Justification: Our framework is a general-purpose explanation algorithm. It is not tailored to any specific application.
    \item[] Guidelines:
    \begin{itemize}
        \item The answer \answerNA{} means that there is no societal impact of the work performed.
        \item If the authors answer \answerNA{} or \answerNo, they should explain why their work has no societal impact or why the paper does not address societal impact.
        \item Examples of negative societal impacts include potential malicious or unintended uses (e.g., disinformation, generating fake profiles, surveillance), fairness considerations (e.g., deployment of technologies that could make decisions that unfairly impact specific groups), privacy considerations, and security considerations.
        \item The conference expects that many papers will be foundational research and not tied to particular applications, let alone deployments. However, if there is a direct path to any negative applications, the authors should point it out. For example, it is legitimate to point out that an improvement in the quality of generative models could be used to generate Deepfakes for disinformation. On the other hand, it is not needed to point out that a generic algorithm for optimizing neural networks could enable people to train models that generate Deepfakes faster.
        \item The authors should consider possible harms that could arise when the technology is being used as intended and functioning correctly, harms that could arise when the technology is being used as intended but gives incorrect results, and harms following from (intentional or unintentional) misuse of the technology.
        \item If there are negative societal impacts, the authors could also discuss possible mitigation strategies (e.g., gated release of models, providing defenses in addition to attacks, mechanisms for monitoring misuse, mechanisms to monitor how a system learns from feedback over time, improving the efficiency and accessibility of ML).
    \end{itemize}
    
\item {\bf Safeguards}
    \item[] Question: Does the paper describe safeguards that have been put in place for responsible release of data or models that have a high risk for misuse (e.g., pre-trained language models, image generators, or scraped datasets)?
    \item[] Answer: \answerNA{} 
    \item[] Justification: We are not aware of any particular risks for misuse of the results presented in this paper.
    \begin{itemize}
        \item The answer \answerNA{} means that the paper poses no such risks.
        \item Released models that have a high risk for misuse or dual-use should be released with necessary safeguards to allow for controlled use of the model, for example by requiring that users adhere to usage guidelines or restrictions to access the model or implementing safety filters. 
        \item Datasets that have been scraped from the Internet could pose safety risks. The authors should describe how they avoided releasing unsafe images.
        \item We recognize that providing effective safeguards is challenging, and many papers do not require this, but we encourage authors to take this into account and make a best faith effort.
    \end{itemize}

\item {\bf Licenses for existing assets}
    \item[] Question: Are the creators or original owners of assets (e.g., code, data, models), used in the paper, properly credited and are the license and terms of use explicitly mentioned and properly respected?
    \item[] Answer: \answerNA{} 
    \item[] Justification: All datasets are from OpenML, and our implementation uses only open-source Python libraries, including \texttt{pytorch} and \texttt{scikit-learn}. We do not use any proprietary third-party code. 
    \item[] Guidelines:
    \begin{itemize}
        \item The answer \answerNA{} means that the paper does not use existing assets.
        \item The authors should cite the original paper that produced the code package or dataset.
        \item The authors should state which version of the asset is used and, if possible, include a URL.
        \item The name of the license (e.g., CC-BY 4.0) should be included for each asset.
        \item For scraped data from a particular source (e.g., website), the copyright and terms of service of that source should be provided.
        \item If assets are released, the license, copyright information, and terms of use in the package should be provided. For popular datasets, \url{paperswithcode.com/datasets} has curated licenses for some datasets. Their licensing guide can help determine the license of a dataset.
        \item For existing datasets that are re-packaged, both the original license and the license of the derived asset (if it has changed) should be provided.
        \item If this information is not available online, the authors are encouraged to reach out to the asset's creators.
    \end{itemize}

\item {\bf New assets}
    \item[] Question: Are new assets introduced in the paper well documented and is the documentation provided alongside the assets?
    \item[] Answer: \answerYes{} 
    \item[] Justification: We provide comments to most of the Python code, and the README file includes instructions for how to use the code and reproduce the main results.
    \item[] Guidelines:
    \begin{itemize}
        \item The answer \answerNA{} means that the paper does not release new assets.
        \item Researchers should communicate the details of the dataset\slash code\slash model as part of their submissions via structured templates. This includes details about training, license, limitations, etc. 
        \item The paper should discuss whether and how consent was obtained from people whose asset is used.
        \item At submission time, remember to anonymize your assets (if applicable). You can either create an anonymized URL or include an anonymized zip file.
    \end{itemize}

\item {\bf Crowdsourcing and research with human subjects}
    \item[] Question: For crowdsourcing experiments and research with human subjects, does the paper include the full text of instructions given to participants and screenshots, if applicable, as well as details about compensation (if any)? 
    \item[] Answer: \answerNA{} 
    \item[] Justification: The paper does not involve crowdsourcing nor research with human subjects.
    \item[] Guidelines:
    \begin{itemize}
        \item The answer \answerNA{} means that the paper does not involve crowdsourcing nor research with human subjects.
        \item Including this information in the supplemental material is fine, but if the main contribution of the paper involves human subjects, then as much detail as possible should be included in the main paper. 
        \item According to the NeurIPS Code of Ethics, workers involved in data collection, curation, or other labor should be paid at least the minimum wage in the country of the data collector. 
    \end{itemize}

\item {\bf Institutional review board (IRB) approvals or equivalent for research with human subjects}
    \item[] Question: Does the paper describe potential risks incurred by study participants, whether such risks were disclosed to the subjects, and whether Institutional Review Board (IRB) approvals (or an equivalent approval/review based on the requirements of your country or institution) were obtained?
    \item[] Answer: \answerNA{} 
    \item[] Justification: The paper does not involve research with human subjects.
    \item[] Guidelines:
    \begin{itemize}
        \item The answer \answerNA{} means that the paper does not involve crowdsourcing nor research with human subjects.
        \item Depending on the country in which research is conducted, IRB approval (or equivalent) may be required for any human subjects research. If you obtained IRB approval, you should clearly state this in the paper. 
        \item We recognize that the procedures for this may vary significantly between institutions and locations, and we expect authors to adhere to the NeurIPS Code of Ethics and the guidelines for their institution. 
        \item For initial submissions, do not include any information that would break anonymity (if applicable), such as the institution conducting the review.
    \end{itemize}

\item {\bf Declaration of LLM usage}
    \item[] Question: Does the paper describe the usage of LLMs if it is an important, original, or non-standard component of the core methods in this research? Note that if the LLM is used only for writing, editing, or formatting purposes and does \emph{not} impact the core methodology, scientific rigor, or originality of the research, declaration is not required.
    \item[] Answer: \answerNA{} 
    \item[] Justification: We did not use LLMs for the main theoretical, methodological, and experimental contributions of this paper. In line with the NeurIPS LLM policy, we only used LLMs for spell-checking, grammar suggestions, and programming assistance.
    \item[] Guidelines:
    \begin{itemize}
        \item The answer \answerNA{} means that the core method development in this research does not involve LLMs as any important, original, or non-standard components.
        \item Please refer to our LLM policy in the NeurIPS handbook for what should or should not be described.
    \end{itemize}

\end{enumerate}

\end{document}